\documentclass[lettersize,journal]{IEEEtran}
\usepackage{amsmath,amsfonts}
\usepackage{algorithmic}
\usepackage{algorithm}
\usepackage{array}
\usepackage[caption=false,font=small]{subfig}
\usepackage{textcomp}
\usepackage{stfloats}
\usepackage{url}
\usepackage{verbatim}
\usepackage{graphicx}
\usepackage{cite}
\usepackage{amssymb}
\usepackage{balance}

\begin{document}

\title{ODD: Omni Differential Drive for Simultaneous Reconfiguration and Omnidirectional Mobility of Wheeled Robots}

\author{Ziqi~Zhao, 
        Peijia~Xie, 
        and Max~Q.-H.~Meng,~\IEEEmembership{Fellow,~IEEE}
\thanks{This work is partially supported by Shenzhen Key Laboratory of Robotics Perception and Intelligence (ZDSYS20200810171800001), Shenzhen Science and Technology Program under Grant RCBS20221008093305007, 20231115141459001, Young Elite Scientists Sponsorship Program by CAST under Grant 2023QNRC001, High level of special funds(G03034K003)from Southern University of Science and Technology, Shenzhen, China. \emph{(Corresponding author: Max~Q.-H.~Meng.)}} 
\thanks{Ziqi Zhao, Peijia Xie and Max Q.-H. Meng are with Shenzhen Key Laboratory of Robotics Perception and Intelligence, and the Department of Electronic and Electrical Engineering, Southern University of Science and Technology, Shenzhen, China.(Email: zhaozq2020@mail.sustech.edu.cn, xiepj2022@mail.sustech.edu.cn, max.meng@ieee.org)}
\thanks{Max Q.-H. Meng is also a Professor Emeritus in the Department of Electronic Engineering at The Chinese University of Hong Kong in Hong Kong and was a Professor in the Department of Electrical and Computer Engineering at the University of Alberta in Canada.}
}


\maketitle

\begin{abstract}
Wheeled robots are highly efficient in human living environments. However, conventional wheeled designs, with their limited degrees of freedom and constraints in robot configuration, struggle to simultaneously achieve stability, passability, and agility due to varying footprint needs. This paper proposes a novel robot drive model inspired by human movements, termed as the Omni Differential Drive (ODD). The ODD model innovatively utilizes a lateral differential drive to adjust wheel spacing without adding additional actuators to the existing omnidirectional drive. This approach enables wheeled robots to achieve both simultaneous reconfiguration and omnidirectional mobility. To validate the feasibility of the ODD model, a functional prototype was developed, followed by comprehensive kinematic analyses. Control systems for self-balancing and motion control were designed and implemented. Experimental validations confirmed the feasibility of the ODD mechanism and the effectiveness of the control strategies. The results underline the potential of this innovative drive system to enhance the mobility and adaptability of robotic platforms.
\end{abstract}


\section{Introduction}
\IEEEPARstart{T}{he} rapid advancement of robotics technology has led to its widespread application in many areas of human life. Wheeled robots have distinct advantages and disadvantages compared to legged robots. Wheeled robots generally provide better efficiency and speed on smooth surfaces, while legged robots are better adapted for complex and uneven terrains\cite{raj_comprehensive_2022,rubio_review_2019,bigdog_2023}. However, most environments where mobile robots are used in human life involve smooth surfaces. For instance, guidance robots, disinfection robots, cleaning robots, and delivery robots primarily operate in hotels, restaurants, airports, office buildings, and residential settings. These applications underscore the importance of optimizing wheeled robot designs for such environments.

Enhancing wheeled robot performance requires balancing three critical properties: passability, agility, and stability. Passability is the ability to navigate through narrow spaces, requiring a small footprint. agility is the capability of omnidirectional movement, allowing the robot to maneuver freely in any direction. Stability requires having a larger footprint to maintain balance and prevent tipping. However, these three properties often conflict. Designing a system that ensures passability, agility, and stability simultaneously is a significant challenge in wheeled robot development.

Static stability is typically achieved through the support polygon, defined by the contact points of the robot's wheels on the ground. The size and shape of this support polygon determine the robot's static stability\cite{vukobratovic_zero-moment_2004}. However, increasing the footprint to enlarge the support polygon can negatively affect the robot's passability in narrow spaces. This issue is often addressed through reconfiguration, allowing dynamic adjustment of the footprint size. The footprint can be enlarged to enhance static stability when needed, or reduced to improve passability in tighter spaces or when crossing obstacles.

Wheel-legged robots adjust the support polygon using leg joints. Yun et al. proposed a wheel-legged robot that alters the support polygon using leg joints while maintaining the orientation of the four Mecanum wheels via a parallel link mechanism in the legs\cite{yun_development_2021}. Li et al. introduced a multi-mode robot that adjusts the support polygon size via leg joints and switches to two collinear Mecanum wheels configurations to reduce the footprint\cite{li_DesignControlTransformable_2024}. Pankert et al. proposed a reconfigurable robot base using joint mechanisms similar to legs, allowing the robot to change the position of the four wheels without altering the height, thus adjusting the support polygon shape\cite{pankert_design_2022}. Some studies add degrees of freedom specifically to adjust the wheel spacing and change the support polygon size. Fuchs et al. presented a mobile platform for a humanoid upper body, named Rollin’ Justin, which adjusts the wheel positions and changes the support polygon shape and size through the driving force of the four steering wheels\cite{fuchs_rollin_2009}. Karamipour et al. proposed a mobile robot with four Mecanum wheels and transverse prismatic joints between the wheels on each side to adjust the form through forces generated by the Mecanum wheels' rotation\cite{karamipour_omnidirectional_2019}. Additionally, Karamipour et al. proposed a mobile robot with four omni-wheels, adding a linear actuator between the wheels on each side to adjust the spacing\cite{karamipour_reconfigurable_2020}. Furthermore, the height is adjusted using a linear actuator to meet different requirements. Both of Karamipour et al.’s works were validated only in simulations and lacked mechanical implementation. Bai et al. designed a lateral deformation tracked robot using a three-stage telescopic structure to adjust the spacing between two tracked modules\cite{bai_design_2022}. However, the lateral movement of the tracked modules must overcome the sliding friction between the tracks and the ground.

Passability and agility can be achieved by reducing the footprint and enabling omnidirectional movement. A support polygon requires at least three non-collinear support points. To achieve a smaller footprint, self-balancing control is typically used to maintain dynamic stability, allowing fewer than three support points or collinear support points. Ball-wheel robots can achieve omnidirectional movement with single-point support \cite{lauwers_dynamically_2006,chen_design_2013,minniti_whole-body_2019,kumagai_development_2008}. Additionally, single omnidirectional wheels add driven rollers to the conventional wheel's rolling degree of freedom, enabling lateral movement \cite{shen_omburo_2020}. Systems with single-point support tend to oscillate when subjected to external disturbances and cannot remain stationary, making them highly demanding in terms of driving torque and energy consumption. In contrast, configurations with two-point support or multiple collinear support points achieve dynamic stability in one direction through self-balancing while maintaining static stability in another direction. Examples include the dual-ball mechanism \cite{gao_dual_ball_2022} and the collinear Mecanum wheel approach \cite{Reynolds-Haertle2011,miyakoshi_omnidirectional_2017,watson_collinear_2021}. The strength of static stability depends on the distance between contact points.

\begin{figure}[!t]
    \centering
    \includegraphics[width=3.3in]{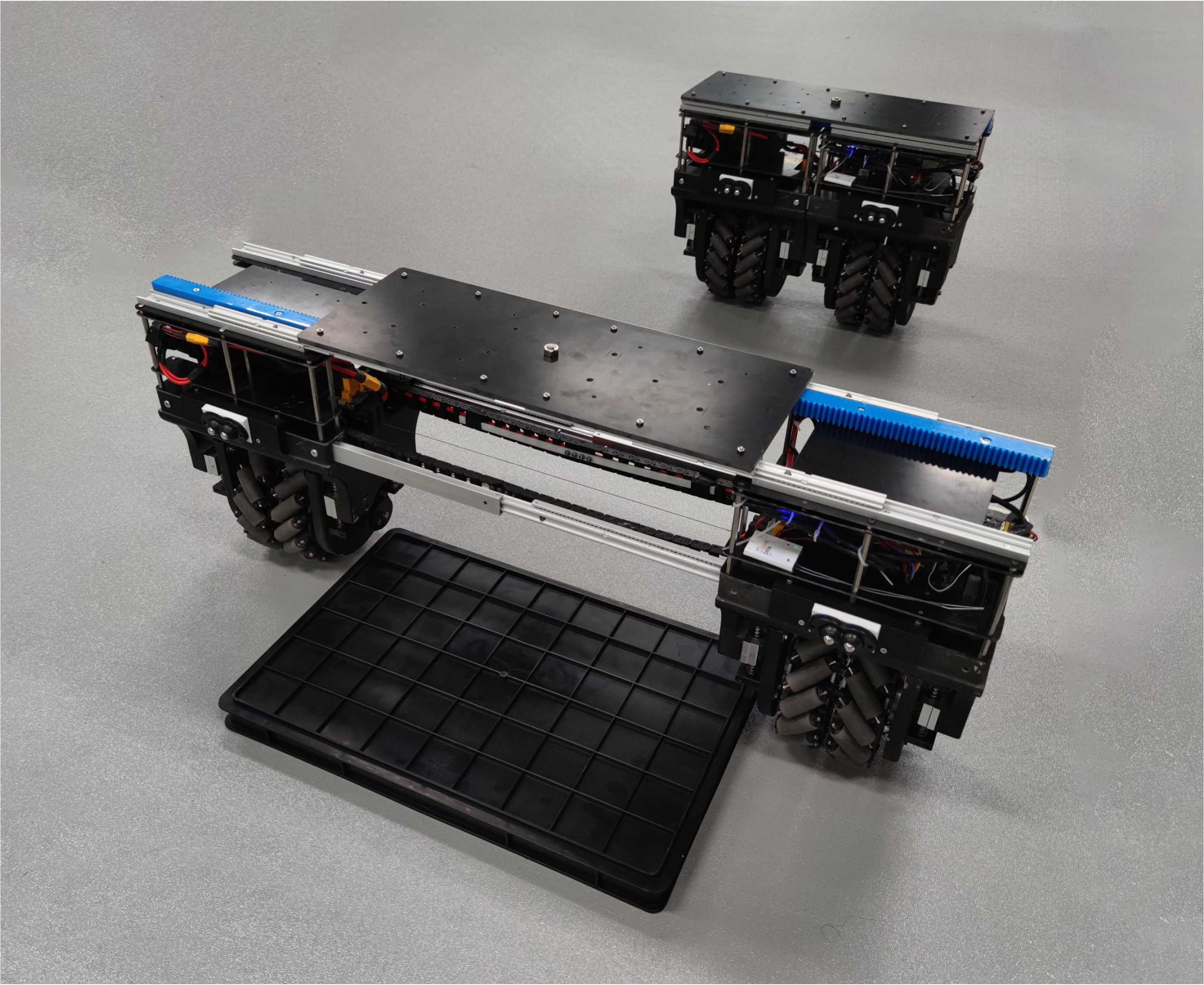}
    \caption{Proposed Prototype which can simultaneous reconfigure and omnidirectional mobile using the Omni Differential Drive (ODD).}
    \label{fig_home}
\end{figure}

Considering these benefits and drawbacks of the mobility method, and inspired by the locomotion of humans and biped robots, we proposed the ODD. This mobility method enables omnidirectional movement and reconfiguration of wheeled mobile platforms, simultaneously meeting the requirements for passability, agility, and stability. To validate this mobility method, we designed a prototype, as shown in Fig. \ref{fig_home}. The primary contributions of this study are summarized as follows:

\begin{enumerate}
\item {A novel Omni Differential Drive (ODD) wheeled mobile model was proposed, enabling the robot to achieve simultaneous omnidirectional movement and reconfiguration, meeting the requirements for mobility, agility, and stability in human living environments.}
\item {A prototype based on a collinear Mecanum wheel mechanism was developed to implement the ODD model, with its kinematics modeled.}
\item {Controllers were designed and implemented to operate the prototype, validating the kinematics of both the ODD model and the prototype.}
\end{enumerate}

The remainder of this paper is organized as follows: Section II introduces the concept. Section III presents the Omni Differential Drive. Section IV describes the prototype design. Section V discusses the modeling and control of the prototype. Section VI details the experimentation and validation. Finally, Section VII concludes the paper and suggests directions for future research.

\section{Concept}

\subsection{Inspired by Human Movements}

Humans, with the extensive degrees of freedom provided by their legs, can achieve omnidirectional movement and flexibly change their direction and orientation. They can turn sideways to navigate narrow passages, such as in kitchens, restaurants, or crowded areas. By increasing the distance between their feet in any direction, they can enhance stability, crucial when dealing with external forces during activities like boxing and Kung Fu, or when experiencing tilting and shaking in transportation modes like airplanes and trains. They can also step over small obstacles, such as scattered items on the floor or puddles on the road. Thus, humans possess excellent passability, stability, and agility.

\begin{figure}[!t]
    \centering
    \subfloat[]{\includegraphics[height=1.5in]{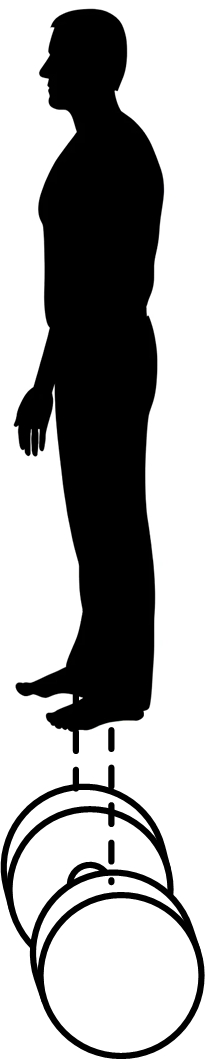}%
    \label{fig_concept_side}}
    \hspace{2pt}
    \subfloat[]{\includegraphics[height=1.5in]{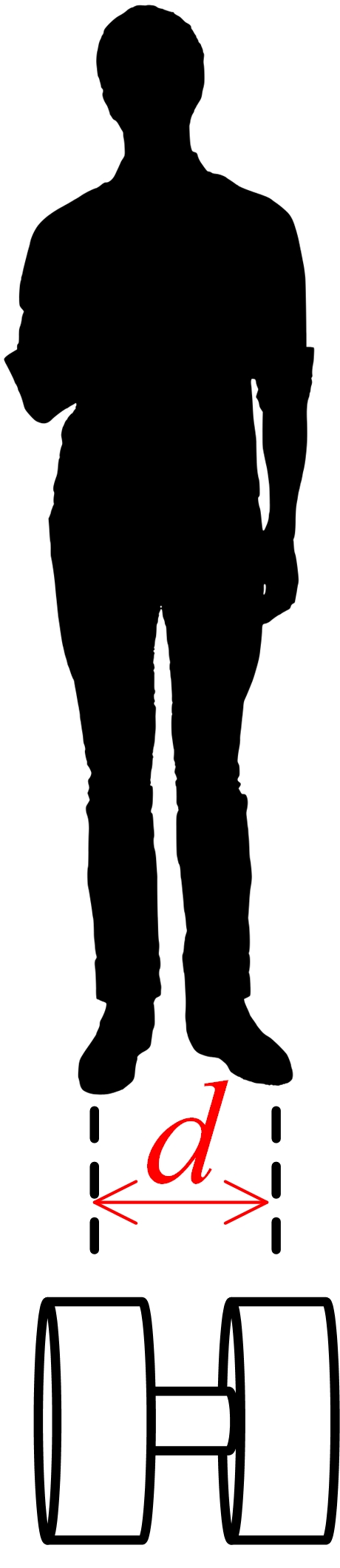}%
    \label{fig_concept_front}}
    \hspace{2pt}
    \subfloat[]{\includegraphics[height=1.5in]{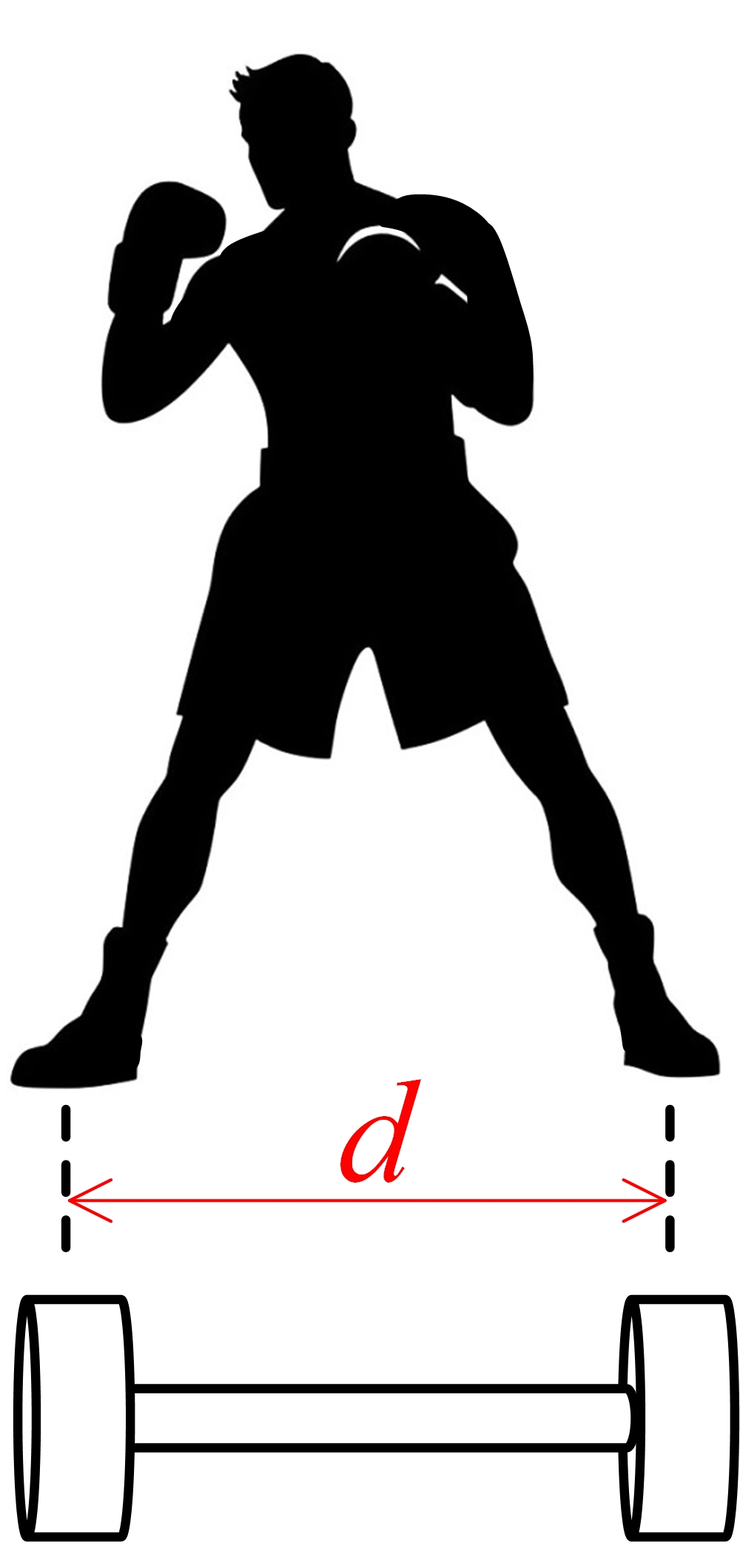}%
    \label{fig_concept_boxing}}
    \hspace{0pt}
    \subfloat[]{\includegraphics[height=1.5in]{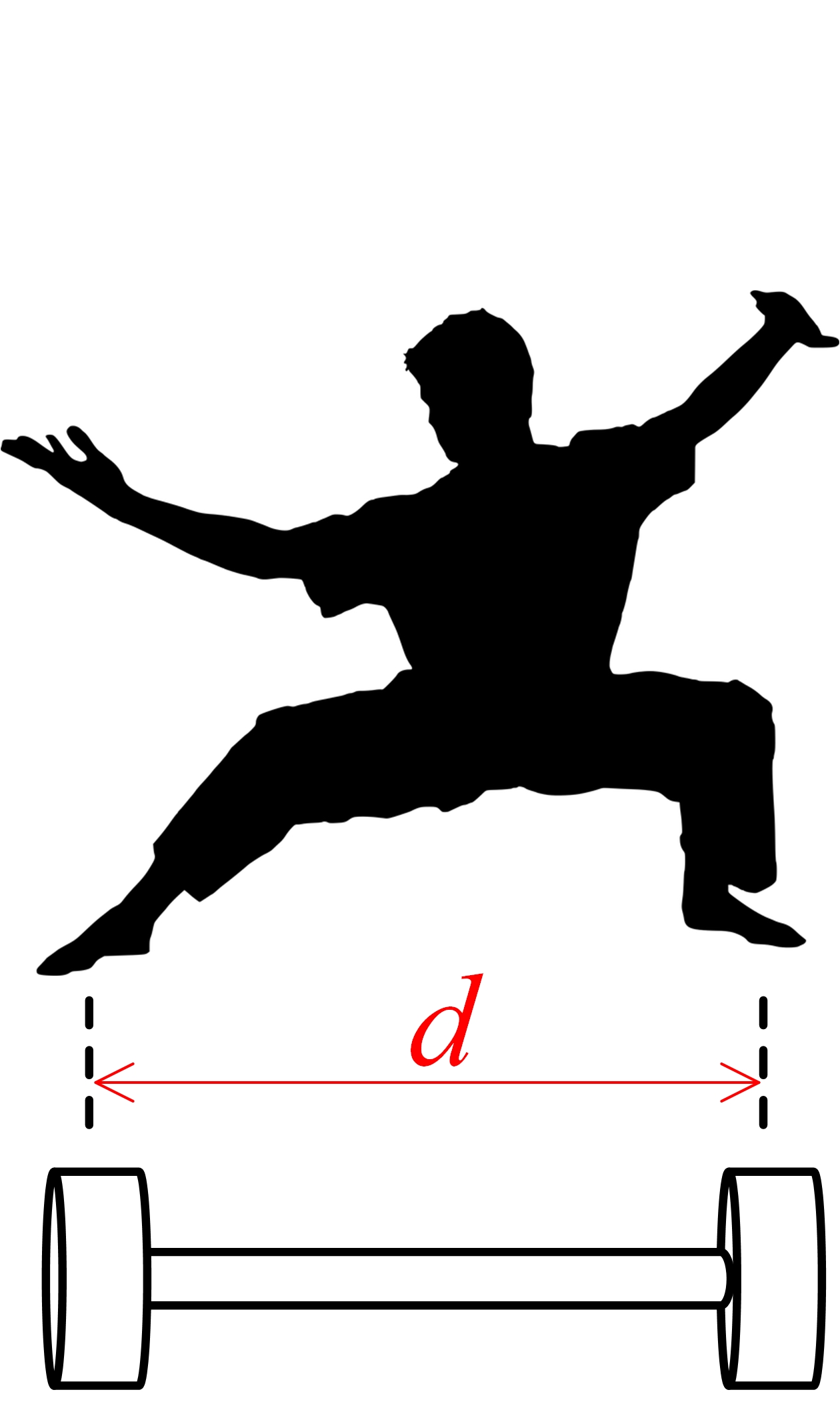}%
    \label{fig_concept_kungfu}}
    \hspace{2pt}
    \subfloat[]{\includegraphics[height=1.5in]{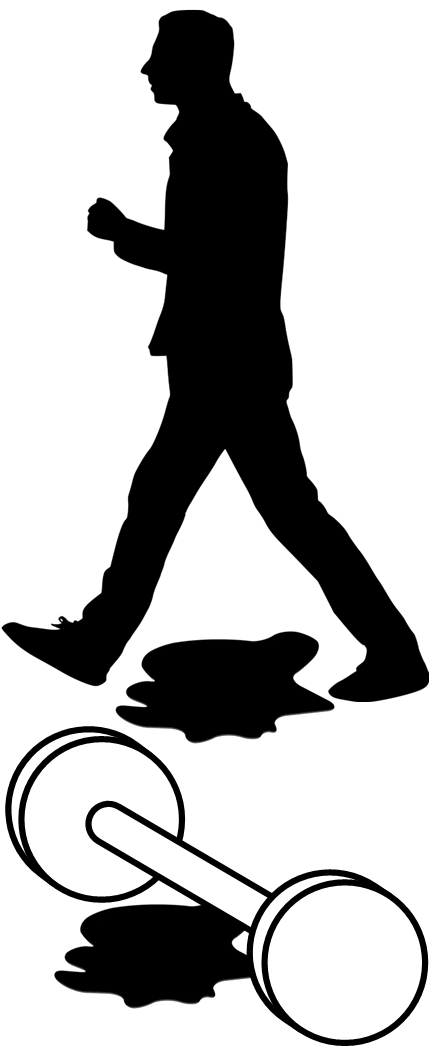}%
    \label{fig_obstacle}}
    \caption{Analogy between human movements and wheeled mobility. (a) Side-view standing or lateral walking. (b) Front-view Standing or longitudinal walking. (c) Boxing. (d) Kung Fu. (e) Obstacle crossing.}
    \label{fig_concept}
\end{figure}

By drawing an analogy between human movement and wheeled mobility, the distance between a human's feet is similar to the wheel spacing of a wheeled platform, as shown in Fig. \ref{fig_concept}. In terms of passability, a smaller wheel spacing results in a smaller footprint, while a larger wheel spacing meets the requirements for overcoming obstacles. Omnidirectional movement allows sideways navigation through narrow passages. Regarding stability, a larger wheel spacing provides greater anti-overturning torque, maintaining static stability. Additionally, omnidirectional movement enables the mobile platform to achieve dynamic stability in all directions. In terms of agility, omnidirectional movement allows rapid changes in direction and orientation. Thus, to satisfy all three properties, the mobile platform must be reconfigurable for changes in wheel spacing and capable of omnidirectional movement.

\subsection{Robotics Applications}

The ODD method meets the requirements for both omnidirectional movement and reconfiguration, making it applicable to various mobile platforms. For example, replacing the driven wheels in a common two-wheel drive platform with two caster wheels and removing the wheel spacing constraint transforms it into a four-wheel mobile platform capable of changing its body width. Similarly, replacing the drive wheels of a two-wheeled self-balancing vehicle with omnidirectional wheels and removing the wheel spacing constraint allows the ODD to achieve omnidirectional movement and reconfiguration. This drive method can also be applied to wheel-legged robots, scooters, and roller skates.

\section{Omni Differential Drive}

\begin{figure}[!t]
    \centering
    \subfloat[]{\includegraphics[height=1.15in]{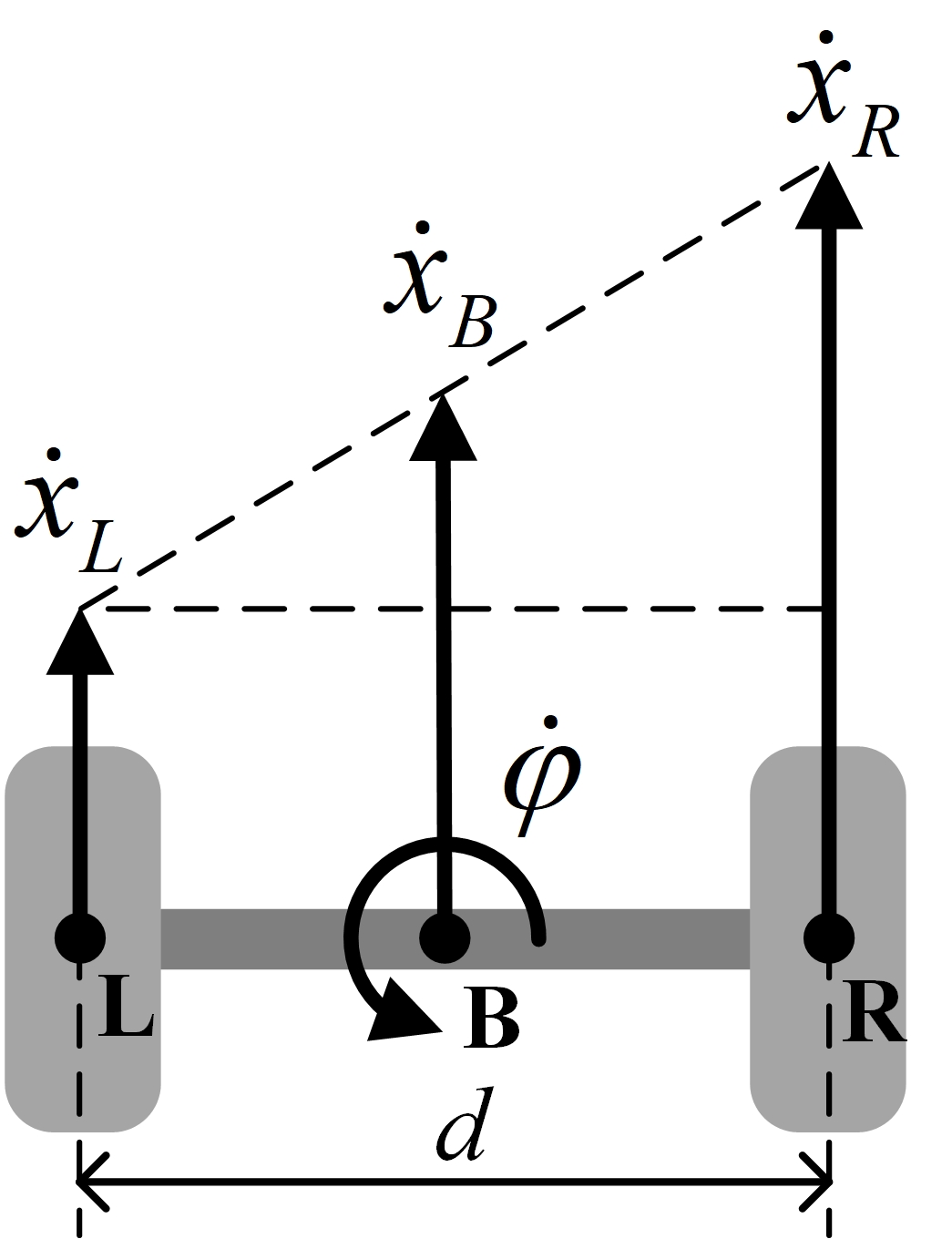}%
    \label{fig_DD}}
    \hspace{0pt}
    \subfloat[]{\includegraphics[height=1.15in]{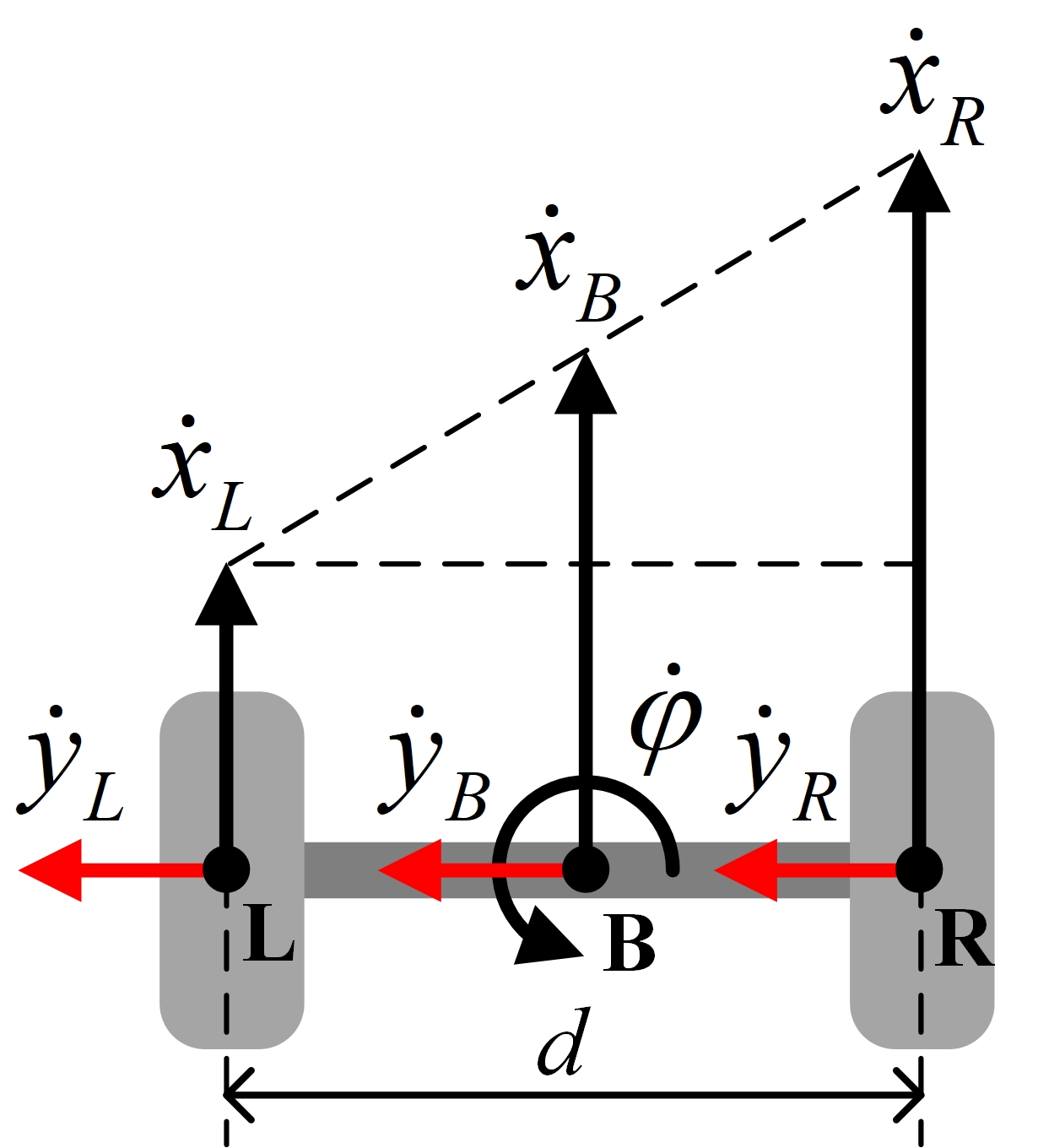}%
    \label{fig_OD}}
    \hspace{0pt}
    \subfloat[]{\includegraphics[height=1.15in]{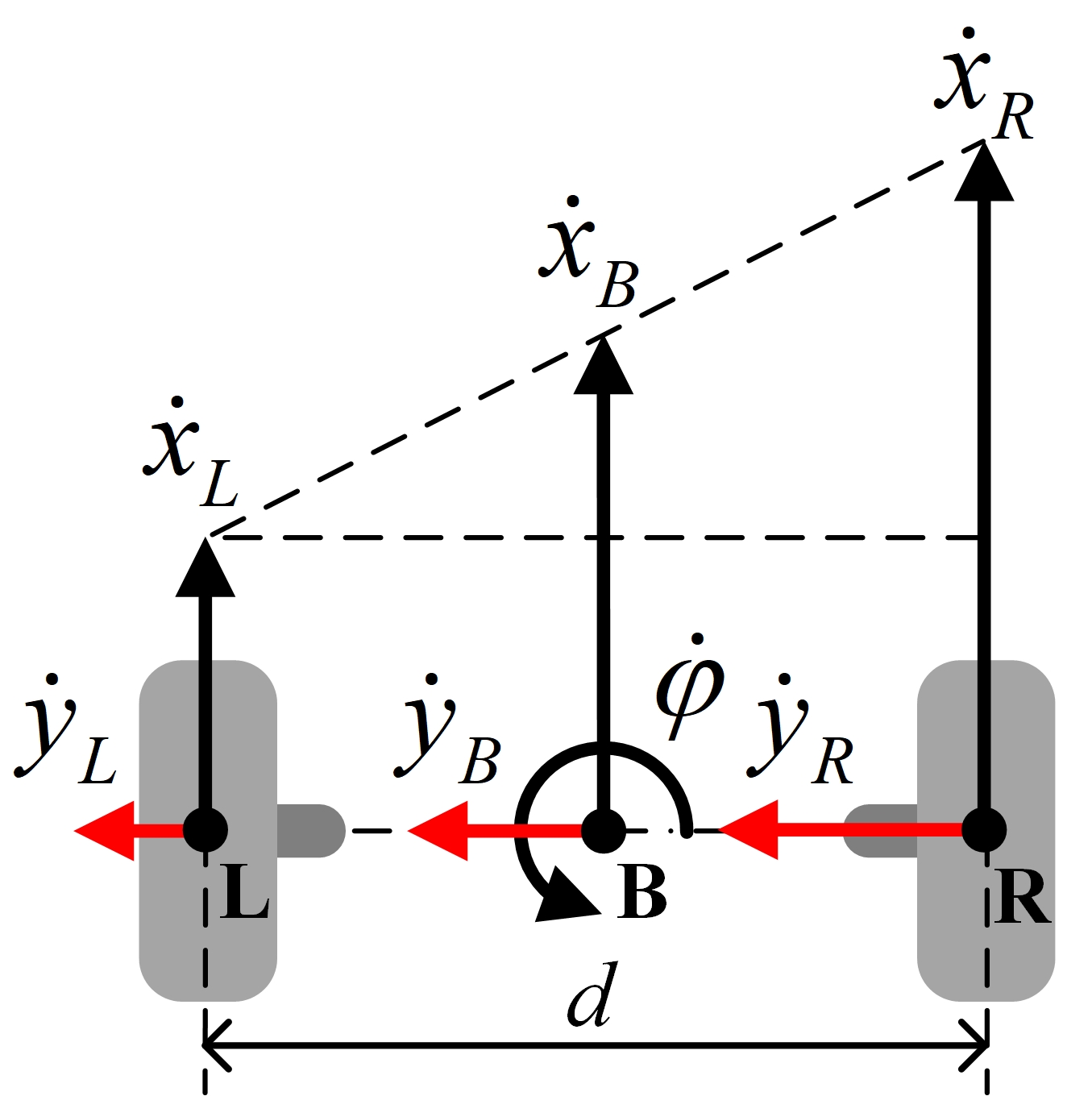}%
    \label{fig_ODD}}
    \caption{Models of drive methods. (a) Differential Drive (DD). (b) Omnidirectional Drive (OD). (c) Proposed Omni Differential Drive (ODD).}
    \label{fig_dd}
\end{figure}
 
Traditional differential drive consists of two single-degree-of-freedom wheels, as shown in Fig. \ref{fig_dd}\subref{fig_DD}. By controlling the linear velocities of the left and right wheels, $\dot{x}_L$ and $\dot{x}_R$, the linear velocity $\dot{x}_B$ and angular velocity $\dot{\varphi}_B$ of the center of the mobile platform can be controlled. The forward and inverse kinematics of the traditional differential drive are

\begin{equation}
    \label{eqn_DD_1}
    \begin{bmatrix}
    \dot{x}_B \\
    \dot{\varphi}_B
    \end{bmatrix}=\begin{bmatrix}
    1/2 & 1/2 \\
    -1/d & 1/d
    \end{bmatrix}\begin{bmatrix}
    \dot{x}_L \\
    \dot{x}_R
    \end{bmatrix},
\end{equation}

\begin{equation}
    \label{eqn_DD_2}
    \begin{bmatrix}
    \dot{x}_L \\
    \dot{x}_R
    \end{bmatrix}=\begin{bmatrix}
    1 & -d/2 \\
    1 & d/2
    \end{bmatrix}\begin{bmatrix}
    \dot{x}_B \\
    \dot{\varphi}_B
    \end{bmatrix}.
\end{equation}

The traditional differential drive model is limited in lateral movement because conventional wheels have only one degree of freedom in rolling. Replacing the differential drive wheels with omnidirectional wheels, which allow both longitudinal and lateral movement, as shown in Fig. \ref{fig_dd}\subref{fig_OD}, makes it possible to control the linear velocities $\dot{x}_B$ and $\dot{y}_B$, and the angular velocity $\dot{\varphi}_B$ of the center of the mobile platform through $\dot{x}_L$, $\dot{y}_L$, $\dot{x}_R$, and $\dot{y}_R$. This enables the robot to achieve omnidirectional mobility. The mechanisms developed by some researchers fit this model\cite{watson_collinear_2021, gao_dual_ball_2022}. The forward and inverse kinematics of the omnidirectional drive model are

\begin{equation}
    \label{eqn_OD_1}
    \begin{bmatrix}
    \dot{x}_B \\
    \dot{y}_B \\
    \dot{\varphi}_B
    \end{bmatrix}=\begin{bmatrix}
    1/2 & 0 & 1/2 & 0 \\
    0 & 1/2 & 0 & 1/2 \\
    -1/d & 0 & 1/d & 0
    \end{bmatrix}\begin{bmatrix}
    \dot{x}_L \\
    \dot{y}_L \\
    \dot{x}_R \\
    \dot{y}_R
    \end{bmatrix},
\end{equation}

\begin{equation}
    \label{eqn_OD_2}
    \begin{bmatrix}
    \dot{x}_L \\
    \dot{y}_L \\
    \dot{x}_R \\
    \dot{y}_R
    \end{bmatrix}=\begin{bmatrix}
    1 & 0 & -d/2 \\
    0 & 1 & 0 \\
    1 & 0 & d/2\\
    0 & 1 & 0
    \end{bmatrix}\begin{bmatrix}
    \dot{x}_B \\
    \dot{y}_B \\
    \dot{\varphi}_B
    \end{bmatrix}.
\end{equation}

As shown in (\ref{eqn_OD_1}) and (\ref{eqn_OD_2}), this system is over-actuated, requiring $\dot{x}_L$ and $\dot{x}_R$ to be equal. Since the wheel spacing $d$ is fixed, any difference between $\dot{x}_L$ and $\dot{x}_R$ would cause slipping or even loss of control. Therefore, this paper proposes ODD based on OD, where the wheel spacing $d$ is variable. By utilizing the difference between $\dot{x}_L$ and $\dot{x}_R$, the change in $d$ can be controlled, as shown in Fig. \ref{fig_dd}\subref{fig_ODD}. This drive model allows for the control of $\dot{x}_B$, $\dot{y}_B$, and $\dot{\varphi}_B$ of the platform's center, as well as the rate of change $\dot{d}$ of $d$ through $\dot{x}_L$, $\dot{y}_L$, $\dot{x}_R$, and $\dot{y}_R$. The forward and inverse kinematics are

\begin{equation}
    \label{eqn_ODD_1}
    \begin{bmatrix}
    \dot{x}_B \\
    \dot{y}_B \\
    \dot{\varphi}_B \\
    \dot{d}
    \end{bmatrix}=\begin{bmatrix}
    1/2 & 0 & 1/2 & 0 \\ 
    0 & 1/2 & 0 & 1/2 \\   
    -1/d & 0 & 1/d & 0 \\
    0 & 1 & 0 & -1
    \end{bmatrix}\begin{bmatrix}
    \dot{x}_L \\
    \dot{y}_L \\
    \dot{x}_R \\
    \dot{y}_R
    \end{bmatrix},
\end{equation}

\begin{equation}
    \label{eqn_ODD_2}
    \begin{bmatrix}
    \dot{x}_L \\
    \dot{y}_L \\
    \dot{x}_R \\
    \dot{y}_R
    \end{bmatrix}=\begin{bmatrix}
    1 & 0 & -d/2 & 0 \\
    0 & 1 & 0 & 1/2 \\
    1 & 0 & d/2 & 0\\
    0 & 1 & 0 & -1/2
    \end{bmatrix}\begin{bmatrix}
    \dot{x}_B \\
    \dot{y}_B \\
    \dot{\varphi}_B \\
    \dot{d}
    \end{bmatrix}.
\end{equation}

The ODD is fully actuated and, compared to the OD, requires no additional actuators to achieve both omnidirectional movement and reconfiguration.

\section{Prototype Design}
To validate the effectiveness and accuracy of the proposed ODD, a prototype was designed, as shown in Fig. \ref{fig_prototype}. The main components are as follows:

\begin{figure}[t]
    \centering
    \subfloat[]{\includegraphics[width=3.3in]{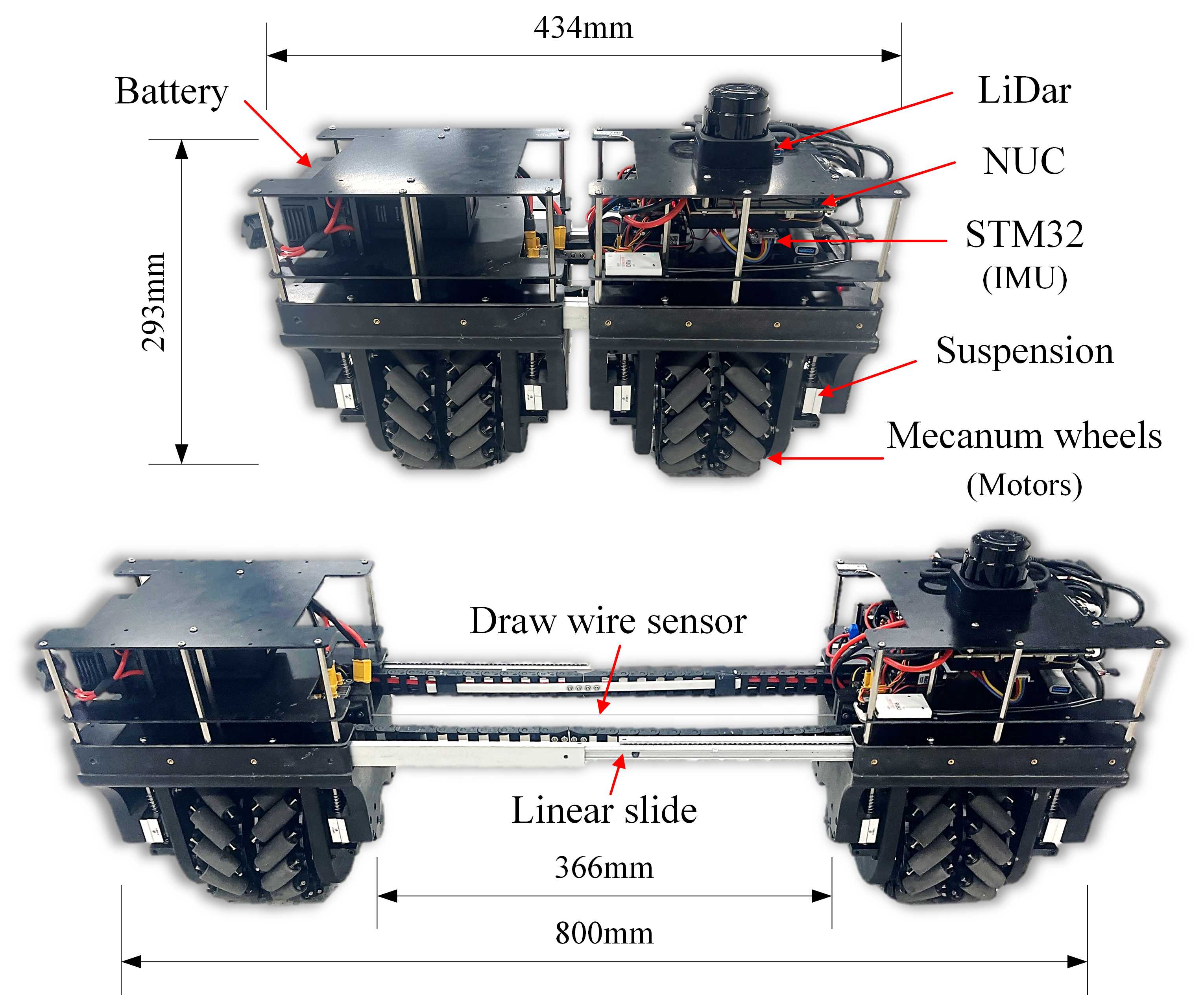}%
    \label{fig_structure}}
    \hspace{0pt}
    \subfloat[]{\includegraphics[width=3.3in]{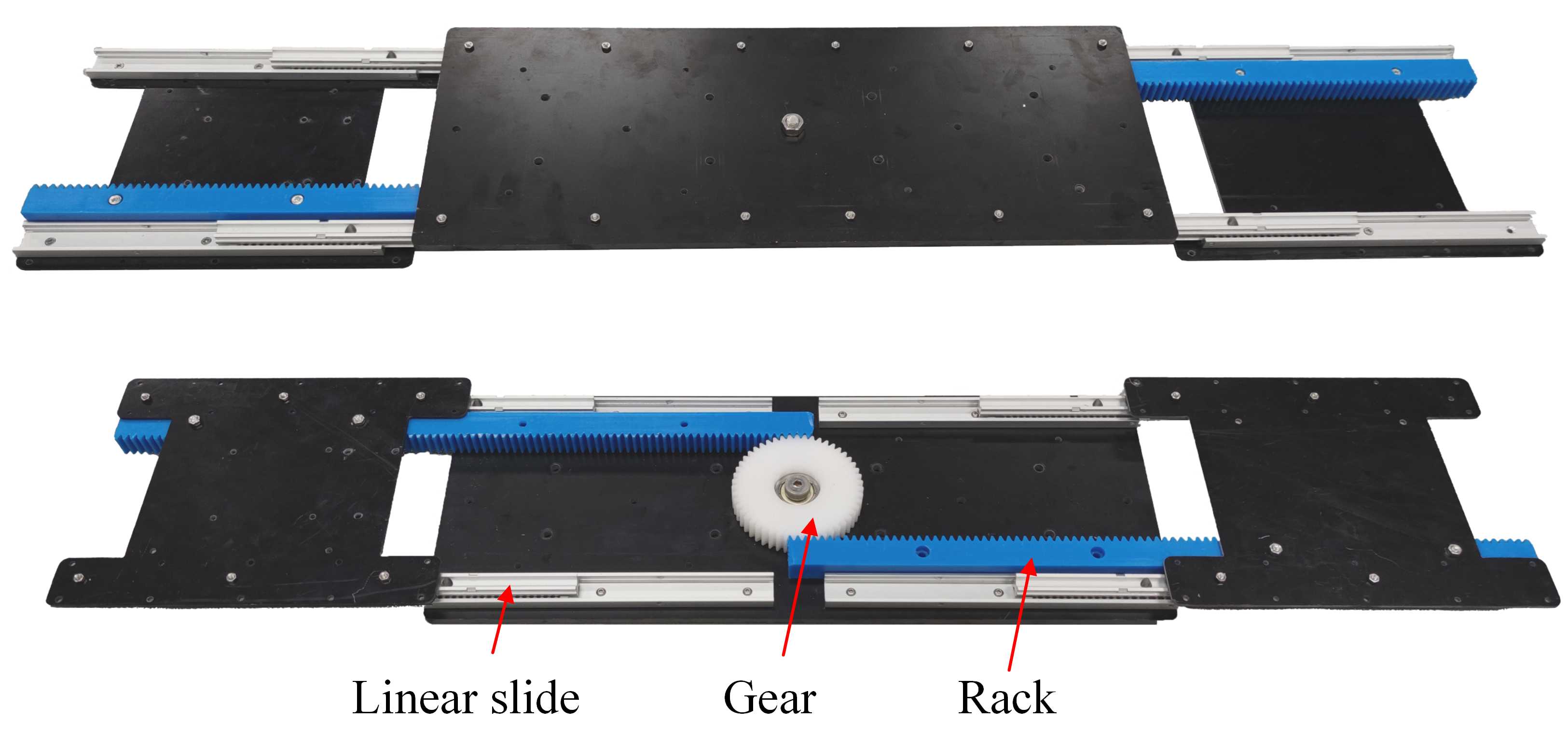}%
    \label{fig_centering}}
    \caption{Components of proposed prototype. (a) Overall structure and dimensions. (b) Self-centering platform.}
    \label{fig_prototype}
\end{figure}

\subsubsection{Active Omnidirectional Wheel}
The active omnidirectional wheel is a crucial component of the ODD. Currently, several active omnidirectional wheel solutions are available, such as collinear Mecanum wheels\cite{watson_collinear_2021}, single-layer omnidirectional wheels\cite{shen_omburo_2020}, and ball wheels\cite{chen_design_2013,gao_dual_ball_2022}. This prototype adopts the collinear Mecanum wheel structure for its simplicity and strong driving force. Each Mecanum wheel is driven by a DJI M3508 DC reduction motor.
\subsubsection{suspension system}
The independent suspension system is designed for each wheel to ensure all wheels maintain contact with the ground even on uneven surfaces.
\subsubsection{Linear slides}
The four collinear Mecanum wheels are divided into two groups, left and right. These two groups are connected by two passive linear slides, ensuring they remain collinear while allowing variable wheel spacing.
\subsubsection{Self-Centering Platform}
Sensors, batteries, computers, and other loads can be placed on the mounting platforms above the left and right wheel groups, as shown in Fig. \ref{fig_prototype}\subref{fig_structure}. Additionally, a self-centering platform can be installed, which remains centered using a rack-and-pinion mechanism, as shown in Fig. \ref{fig_prototype}\subref{fig_centering}.
\subsubsection{Sensors}
A draw-wire sensor is used to measure the wheel spacing with an accuracy of ±0.1\%. An Inertial Measurement Unit (IMU) integrated into the STM32 microcontroller measures angles, angular velocities, and accelerations. A Light Detection and Ranging (LiDAR) sensor is used for global positioning of the robot, and in combination with the measurements from the draw-wire sensor, the pose of the robot's center can be calculated. Additionally, the motor encoders detect the motor's rotation angle and speed, while the feedback current from the motors can be used to calculate motor torque for motion control and balance control.
\subsubsection{Other Components}
The proposed prototype uses an Intel NUC computer as the upper-level controller and an STM32 as the lower-level controller, along with the motor driver boards. Additionally, a 5700mAh battery is included to ensure the robot has sufficient endurance.

\section{Modeling and Control}
\subsection{Kinematics Model}

\begin{figure}[t]
\centering
\includegraphics[width=3in]{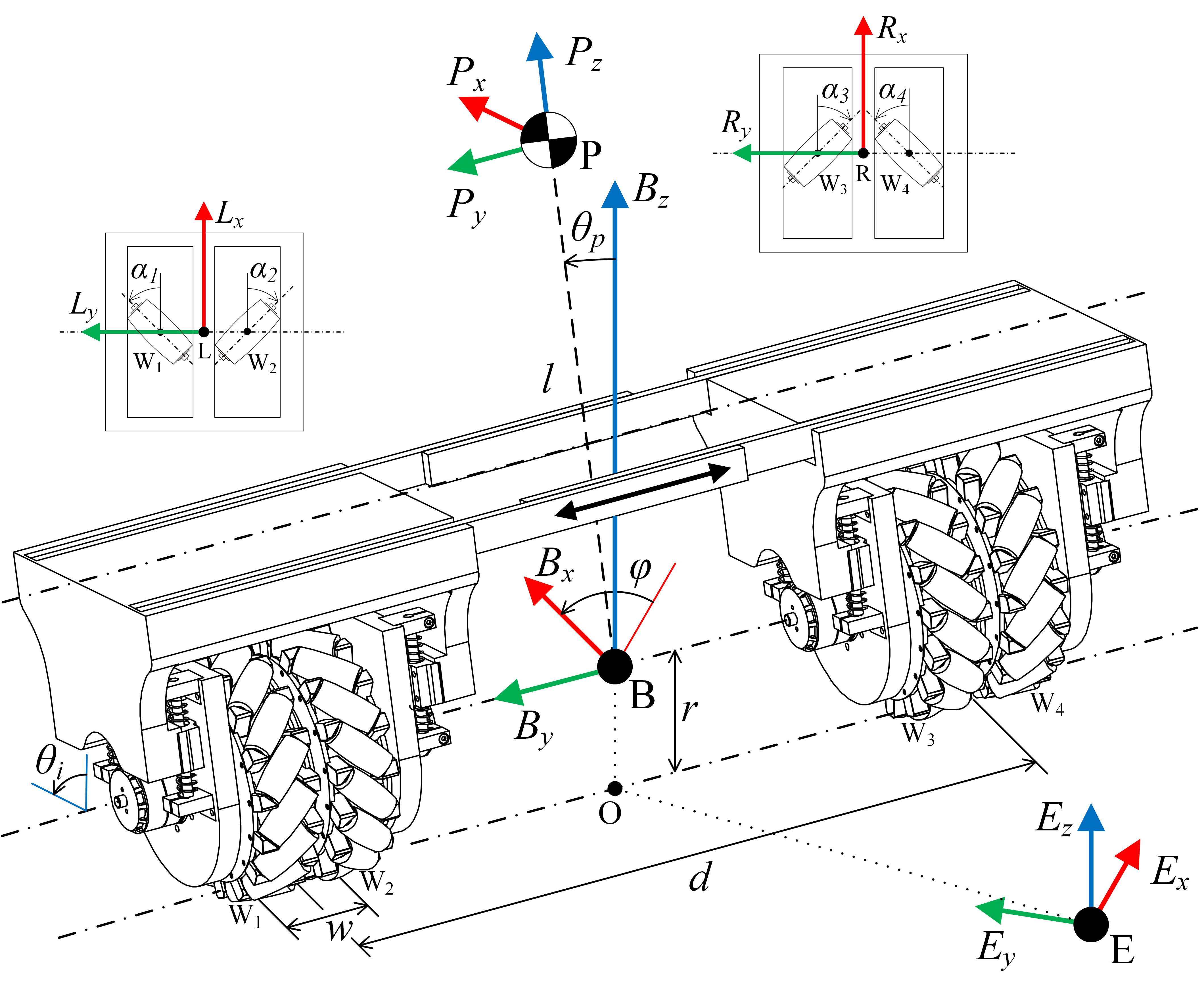}
\caption{Coordinates and parameters for the proposed prototype.}
\label{fig_kinematics}
\end{figure}

First, model the proposed prototype using its coordinates and parameters, as shown in Fig. \ref{fig_kinematics}. The global coordinate system is denoted as $E$, and the moving platform's body coordinate system is denoted as $B$. The origin of coordinate system $B$ is located on the wheel axis, and $B$ is rotated by an angle $\varphi$ around the $z$-axis relative to $E$, i.e., the angle between $E_x$ and $B_x$. The projection of point $B$ onto the $E_{xy}$ plane is point $O$. The four wheels are numbered $i$ from left to right as 1, 2, 3, and 4. The contact points of the wheels with the $E_{xy}$ plane, i.e., the ground, are denoted as $W_i$. The midpoint of $W_1$ and $W_2$ is $L$, and the midpoint of $W_3$ and $W_4$ is $R$. The distance between $L$ and $R$ is $d$. The distance between $W_1$ and $W_4$, and $W_2$ and $W_3$, is $w$. The radius of all four wheels is $r$, and their angular velocities are $\theta_i$. The angle between the roller axis of each Mecanum wheel and the $B_x$ direction is $\alpha_i$.

Typically, modeling Mecanum wheel structures involves directly establishing the kinematic relationships between the four Mecanum wheels and the central point\cite{zimmermann_mecanum_2014,watson_collinear_2021,yun_development_2021}. However, since omnidirectional movement can be achieved through various means, many studies have realized it using different approaches\cite{gao_dual_ball_2022}. In this paper, to demonstrate the universality of the proposed ODD model, we model the prototype using the ODD model. This involves first establishing the motion relationships between points $L$ and $R$ with the four wheels, yielding equations (\ref{eqn_kinematic_L}) and (\ref{eqn_kinematic_R}), respectively, and then deriving the inverse kinematics equation (\ref{eqn_kinematic_LR}). These are substituted into the ODD motion equations (\ref{eqn_ODD_1}) to obtain the kinematic equations (\ref{eqn_kinematic}). In the equations, $T_i$ denotes $\tan{\alpha_i}$.

\begin{equation}
    \label{eqn_kinematic_L}
    \begin{bmatrix}
    \dot{\theta}_1 \\
    \dot{\theta}_2 \\
    \dot{\theta}_3 \\
    \dot{\theta}_4
    \end{bmatrix}=\frac{1}{r}\begin{bmatrix}
    1 & T_1 & -w/2 &  0 \\
    1 & T_2 & w/2 &  0 \\
    1 & T_3 & d-w/2 & - T_3 \\
    1 & T_4 & d+w/2 & - T_4
    \end{bmatrix}\begin{bmatrix}
    \dot{x}_L \\
    \dot{y}_L \\
    \dot{\varphi}_L \\
    \dot{d}
    \end{bmatrix}, 
\end{equation}

\begin{equation}
    \label{eqn_kinematic_R}
    \begin{bmatrix}
    \dot{\theta}_1 \\
    \dot{\theta}_2 \\
    \dot{\theta}_3 \\
    \dot{\theta}_4
    \end{bmatrix}=\frac{1}{r}\begin{bmatrix}
    1 & T_1 & -(d+w/2) &  T_1 \\
    1 & T_2 & -(d-w/2) &  T_2 \\
    1 & T_3 & -w/2 & 0 \\
    1 & T_4 & w/2 & 0
    \end{bmatrix}\begin{bmatrix}
    \dot{x}_R \\
    \dot{y}_R \\
    \dot{\varphi}_R \\
    \dot{d}
    \end{bmatrix}.
\end{equation}

\begin{figure*}[t]
    \begin{equation}
    \label{eqn_kinematic_LR}
    \begin{gathered}
    \begin{bmatrix}
        \dot{x}_L \\
        \dot{y}_L \\
        \dot{x}_R \\
        \dot{y}_R
    \end{bmatrix}
    = \dfrac{r}{2\sigma_1}
    \begin{bmatrix}
        -T_2T_3\sigma_2 + T_2T_4\sigma_3 & T_1T_3\sigma_2 - T_1T_4\sigma_3 & T_4T_1w + T_4T_2w & -T_3T_1w - T_3T_2w \\
        2(T_3d - T_4\sigma_4& - 2( T_3\sigma_5 - T_4d) & -2 T_4w & 2T_3w \\
        -T_2T_3w - T_2T_4w & T_1T_3w + T_1T_4w & -T_4T_1\sigma_3 + T_4T_2\sigma_2 & T_3T_1\sigma_3 - T_3T_2\sigma_2 \\
        2T_2w & - 2T_1w & 2(T_1d - T_2\sigma_5) & - 2( T_1\sigma_4 - T_2d) \\
    \end{bmatrix}
    \begin{bmatrix}
        \dot{\theta}_1 \\
        \dot{\theta}_2 \\
        \dot{\theta}_3 \\
        \dot{\theta}_4
    \end{bmatrix}, 
    \\
    \text{where } \sigma_1 = T_1T_3d - T_1T_4d + T_1T_4w - T_2T_3d - T_2T_3w + T_2T_4d,\ \sigma_2 = 2d + w,\ \sigma_3 = 2d - w,\ \sigma_4 = d - w,\ \sigma_5 = d + w.
    \end{gathered}
    \end{equation}
\end{figure*}

\begin{figure*}[t]
    \begin{equation}
    \label{eqn_kinematic}
    \begin{bmatrix}
        \dot{x}_B \\
        \dot{y}_B \\
        \dot{\varphi}_B \\
        \dot{d}
    \end{bmatrix}
    = \dfrac{r}{2\sigma_1}
    \begin{bmatrix}
        -T_2T_3\sigma_5 + T_2T_4\sigma_4 & T_1T_3\sigma_4 - T_1T_4\sigma_4 & -T_4T_1\sigma_4 + T_4T_2\sigma_4 & T_3T_1\sigma_4 - T_3T_2\sigma_4 \\
        T_2w + T_3d - T_4\sigma_4 & -T_1w - T_3\sigma_4 + T_4d & T_1d - T_2\sigma_4 - T_4w & -T_1\sigma_4 + T_2d + T_3w \\
        2T_2(T_3 - T_4) & 2T_1(-T_3 + T_4) & 2T_4(-T_1 + T_2) & 2T_3(T_1 - T_2) \\
        2(-T_2w + T_3d - T_4\sigma_4) & 2(T_1w - T_3\sigma_5 + T_4d) & 2(-T_1d + T_2\sigma_4 - T_4w) & 2(T_1\sigma_4 - T_2d + T_3w)
    \end{bmatrix}
    \begin{bmatrix}
        \dot{\theta}_1 \\
        \dot{\theta}_2 \\
        \dot{\theta}_3 \\
        \dot{\theta}_4
    \end{bmatrix}. 
    \end{equation}    
\end{figure*}

\subsection{Control Architecture}

\begin{figure}[t]
\centering
\includegraphics[width=3.2in]{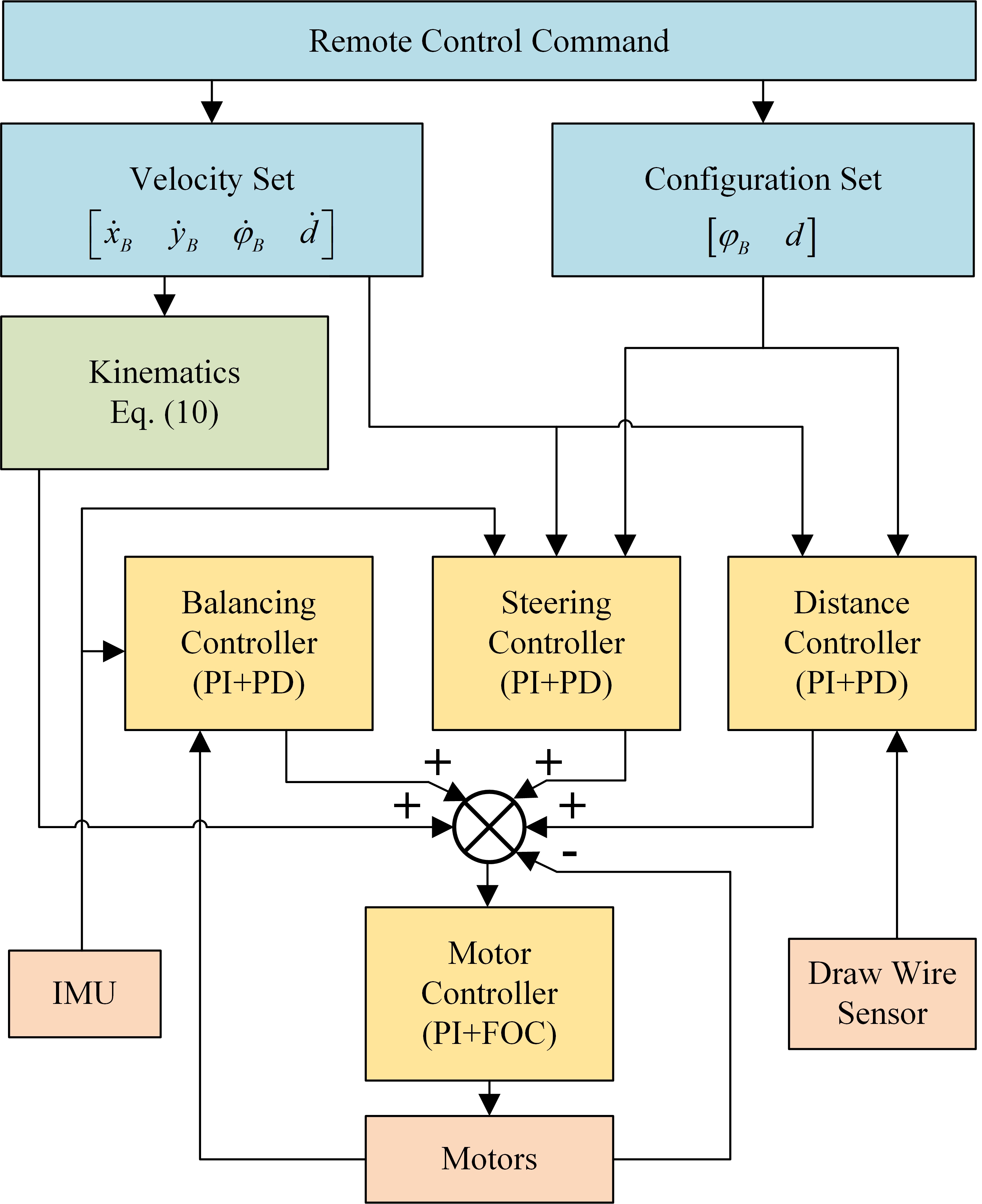}
\caption{Control architecture of the proposed prototype.}
\label{fig_control}
\end{figure}

The frictional force generated by the Mecanum wheel is oriented at a 45-degree angle. By using the distinct distribution of the four wheels, the driving forces in the desired direction of movement are combined, while those in the undesired direction are canceled out. However, due to the collinear arrangement of the Mecanum wheels, additional angular velocity can be generated in certain situations, leading to significant deviations in motion control\cite{li_DesignControlTransformable_2024}. To address this issue, the proposed prototype employs Parallel Cascade Proportional-Integral-Derivative (PID) control to achieve both self-balancing and motion control while minimizing interference from the additional angular velocity. The control architecture is shown in Fig. \ref{fig_control}.

\subsubsection{Balancing Control}
The balancing control uses a cascade PID structure with a position loop Proportional-Derivative (PD) controller and a speed loop Proportional-Integral (PI) controller. The position loop uses the pitch angle from the IMU as feedback; the speed loop uses the time derivative of the wheel motor encoder values as feedback.
\subsubsection{Steering Control}
The steering control uses a cascade PID structure with a position loop PD controller and a speed loop PI controller. The position loop uses the yaw angle from the IMU as feedback; the speed loop uses the time derivative of the yaw angle as feedback.
\subsubsection{Distance Control}
The distance control uses a cascade PID structure with a position loop PD controller and a speed loop PI controller. The position loop uses values from the draw-wire sensor as feedback; the speed loop uses the time derivative of the draw-wire sensor values as feedback.

\subsubsection{Motor Control}
Motor control employs a cascade PID structure with a speed loop PI controller and a current loop PI controller. The speed loop uses the time derivative of the wheel motor encoder values as feedback; the current loop is implemented internally by the DJI C620 Electronic Speed Controller (ESC) using Field-Oriented Control (FOC) to control the motor.

The speed output signals from the balancing control, steering control, and distance control, along with speed commands from the host computer or remote controller, are input into the motor control speed loop. The speed loop processes the combined speed commands and generates current commands through the PID controller. These current commands are then fed into the current loop, which controls the motor current to achieve the desired speed control.

\section{Experiments}
A series of experiments were conducted to validate the omnidirectional mobility, reconfigurability, and passability. These tests confirmed the effectiveness of the proposed prototype and the accuracy of its kinematic model, thereby validating the proposed ODD model. A demonstration video of these tests is included in the supplementary material referenced in the footnotes on the first page of the paper and can be downloaded from the paper’s webpage.

\subsection{Verification of Omnidirectional Mobility}

\begin{figure}
    \centering
    \subfloat[]{\includegraphics[height=1.6in]{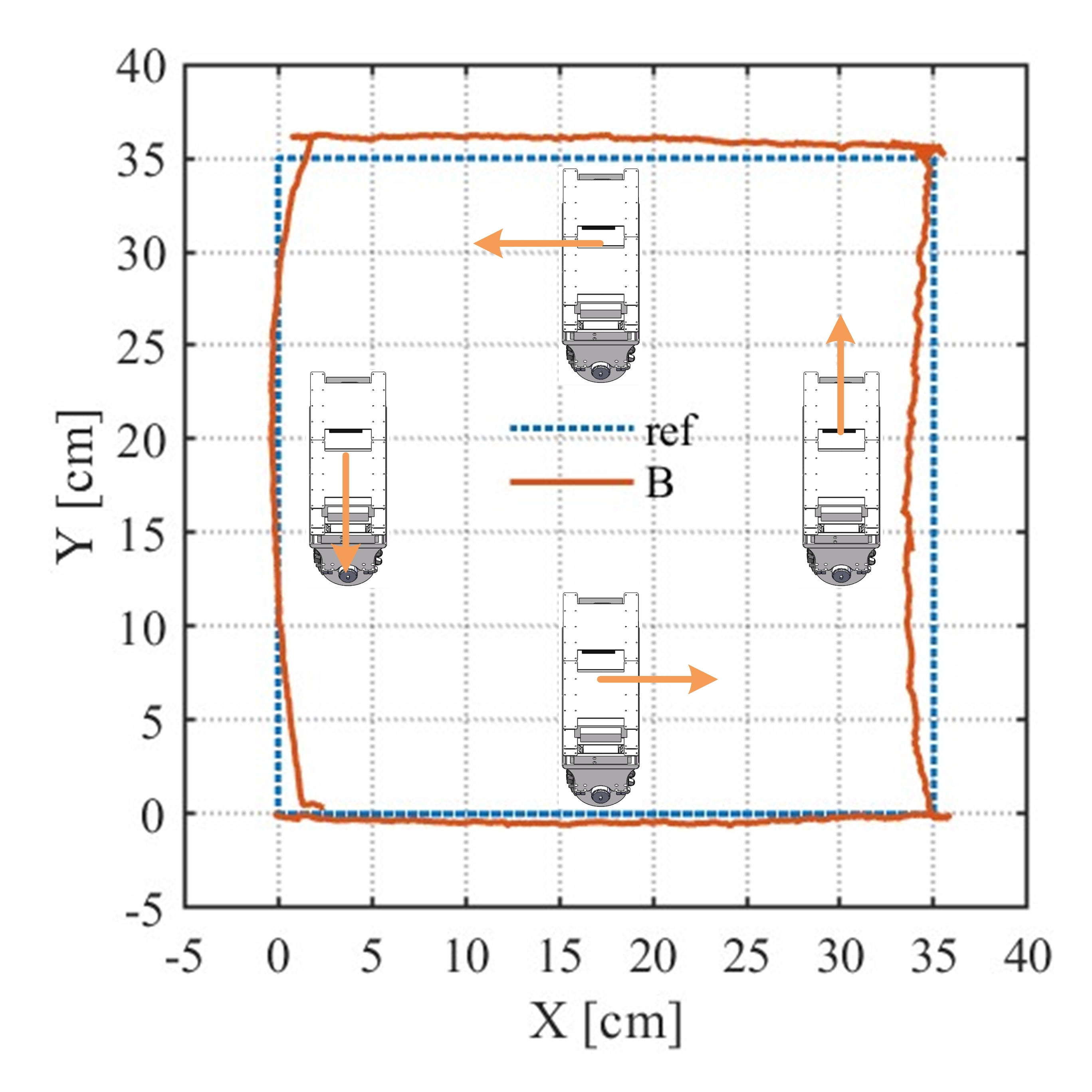}%
    \label{fig_experiment_track_XY_square}}
    \hspace{0pt}
    \subfloat[]{\includegraphics[height=1.6in]{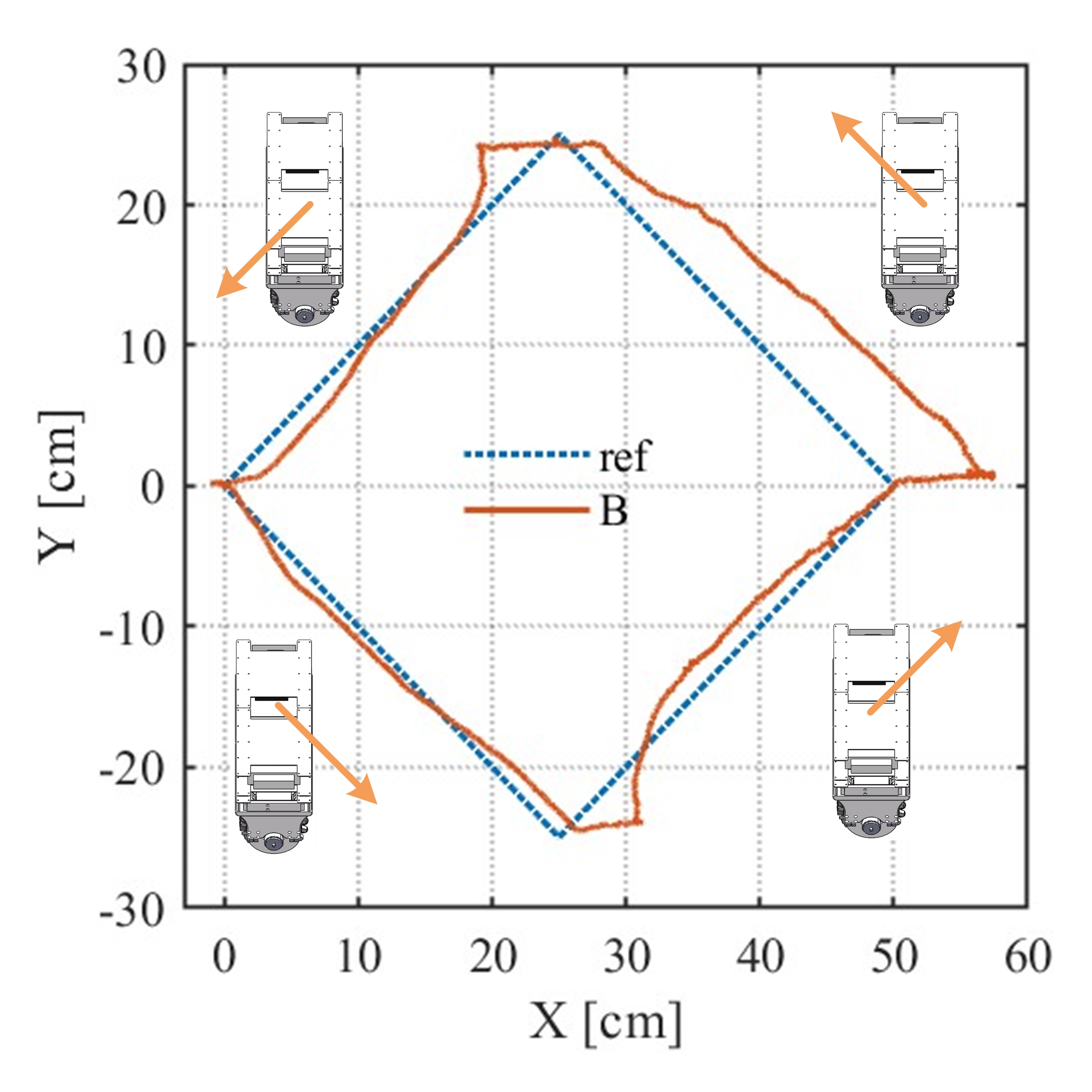}%
    \label{fig_experiment_track_XY_rhombus}}
    \hspace{0pt}
    \subfloat[]{\includegraphics[height=1.6in]{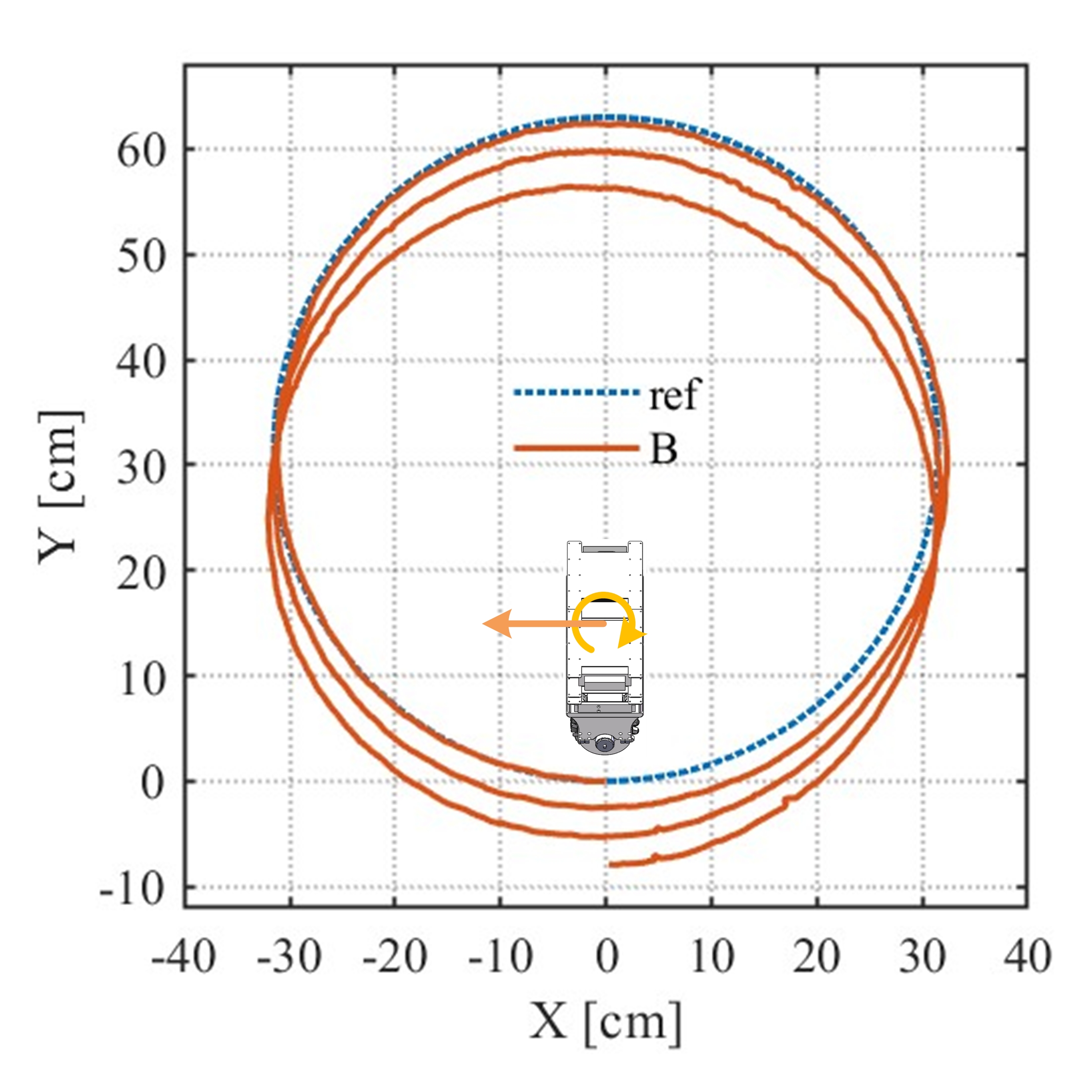}
    \label{fig_experiment_track_XZ}}
    \hspace{0pt}
    \subfloat[]{\includegraphics[height=1.6in]{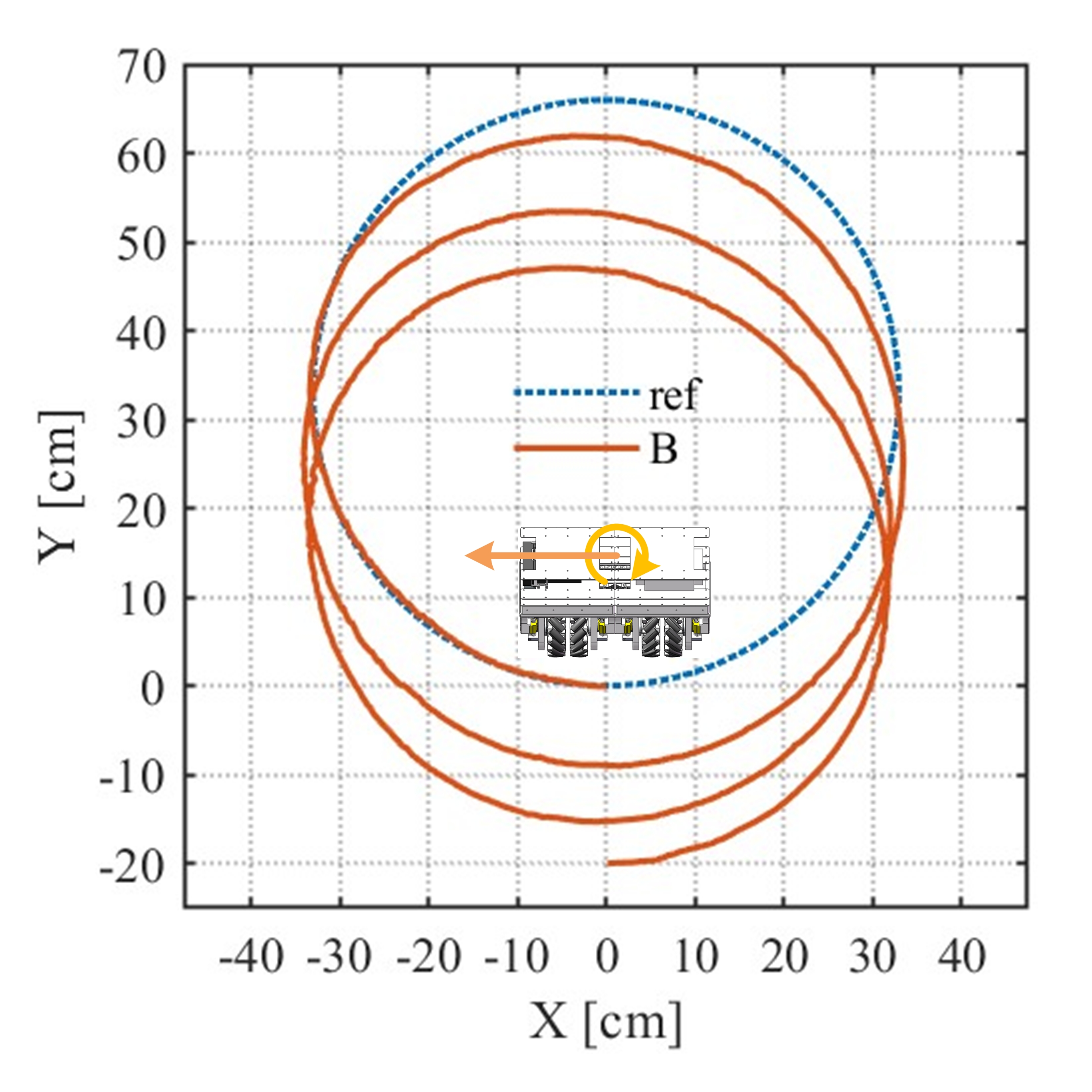}
    \label{fig_experiment_track_YZ}}
    \caption{Paths of center of the prototype. (a) $\dot{x}_B$ or $\dot{y}_B$. (b) $\dot{x}_B$ and $\dot{y}_B$. (C) $\dot{x}_B$ and $\dot{\varphi}_B$. (d) $\dot{y}_B$ and $\dot{\varphi}_B$. The starting coordinates are (0,0).}
    \label{fig_experiment_track_XYZ}
\end{figure}

Omnidirectional movement is achieved through three degrees of freedom: translation along the $B_x$ and $B_y$ axes, and rotation around the $B_z$ axis. Four experiments were designed to verify the accuracy of the prototype's movement according to commands without position compensation. Fig. \ref{fig_experiment_track_XYZ} shows the motion trajectories of the prototype's central point, with data collected by the OptiTrack motion capture system.

In Fig. \ref{fig_experiment_track_XYZ}\subref{fig_experiment_track_XY_square}, velocities $\dot{x}_B$ and $\dot{y}_B$ along the $B_x$ and $B_y$ axes were applied for a given duration $t$. Theoretically, this should produce a square trajectory. The actual trajectory closely matches the theoretical reference, with the starting and ending coordinates nearly overlapping. The observed curvature is mainly due to slight tilting in the forward-backward direction caused by self-balancing along the $B_x$ axis during $B_y$ axis movement.

Fig. \ref{fig_experiment_track_XYZ}\subref{fig_experiment_track_XY_rhombus} shows the result when velocities and durations are applied simultaneously to the $B_x$ and $B_y$ axes, theoretically producing a rhombus trajectory. The actual trajectory differs slightly from the theoretical reference, with the starting and ending coordinates nearly overlapping. The discrepancy is primarily due to the $B_x$ axis's self-balancing control being affected by inertia, resulting in displacement at the four corners of the rhombus. Additionally, the curvature of the trajectory edges is caused by different accelerations along the $B_x$ and $B_y$ axes, leading to curved paths at the beginning and end of each straight segment.

Fig. \ref{fig_experiment_track_XYZ}\subref{fig_experiment_track_XZ} presents the result when velocities along the $B_x$ axis and angular velocities around the $B_z$ axis are applied, theoretically producing a circular trajectory. The actual trajectory closely matches the theoretical reference, but there is a significant difference between the starting and ending coordinates. This deviation is mainly due to the self-balancing control adjusting the $B_x$ axis velocity to maintain balance during movement, thus affecting the overall trajectory.

Fig. \ref{fig_experiment_track_XYZ}\subref{fig_experiment_track_YZ} shows the result when velocities along the $B_y$ axis and angular velocities around the $B_z$ axis are applied, theoretically producing a circular trajectory. The actual trajectory closely matches the theoretical reference, but the deviation between the starting and ending coordinates is larger than that observed in Fig. \ref{fig_experiment_track_XYZ}\subref{fig_experiment_track_XZ}. The primary reasons for this deviation are the effects of self-balancing and the additional angular velocity generated by the collinear arrangement of the Mecanum wheels during $B_y$ axis movement\cite{li_DesignControlTransformable_2024}.

\subsection{Verification of Reconfigurability in Motion}

\begin{figure}
    \centering
    \subfloat[]{\includegraphics[height=1.5in]{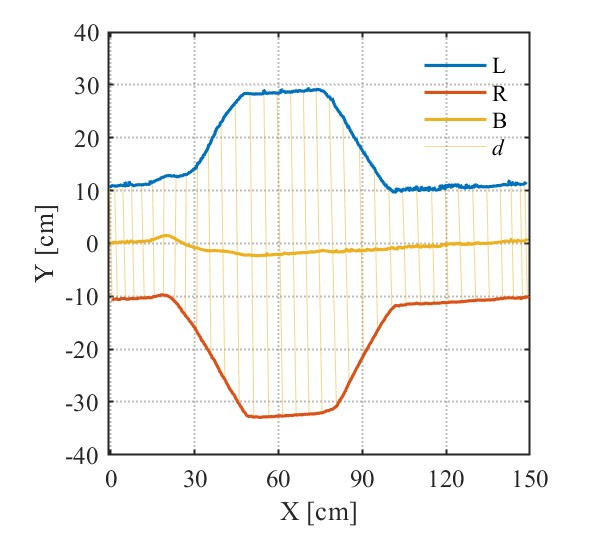}%
    \label{fig_experiment_track_XD_XY}}
    \hspace{0pt}
    \subfloat[]{\includegraphics[height=1.5in]{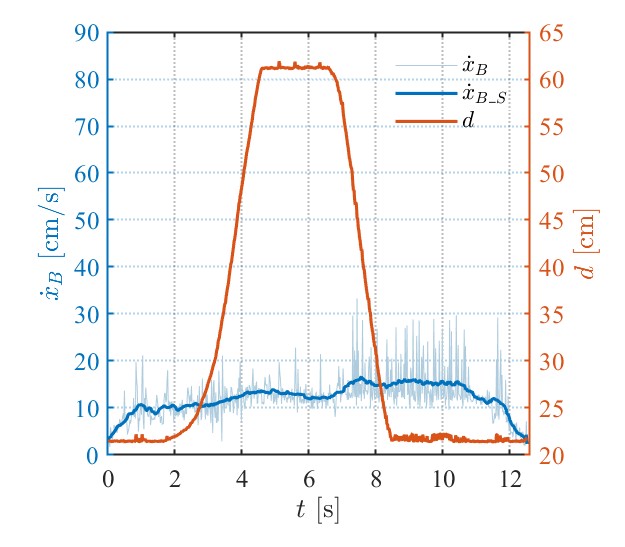}%
    \label{fig_experiment_track_XD_d}}
    \hspace{0pt}
    \subfloat[]{\includegraphics[height=1.5in]{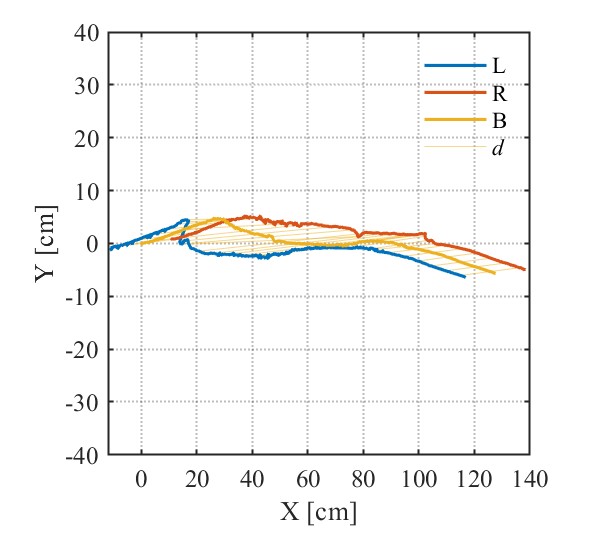}
    \label{fig_experiment_track_YD_XY}}
    \hspace{0pt}
    \subfloat[]{\includegraphics[height=1.5in]{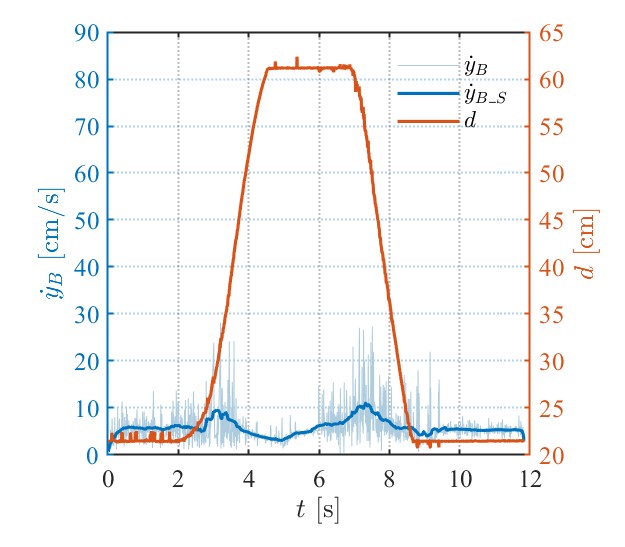}
    \label{fig_experiment_track_YD_d}}
    \hspace{0pt}
    \subfloat[]{\includegraphics[height=1.5in]{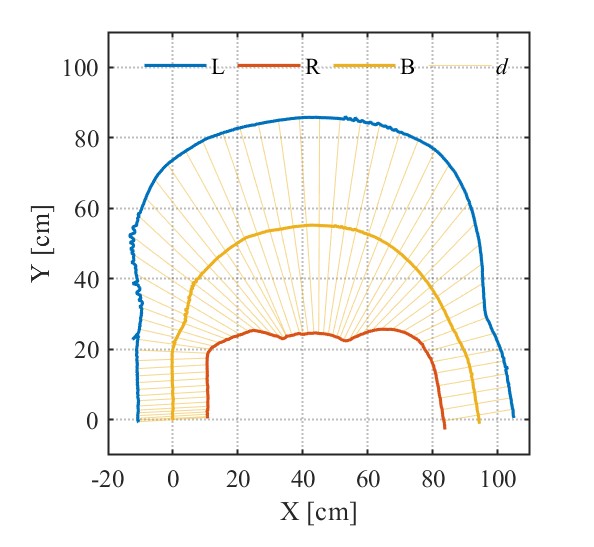}
    \label{fig_experiment_track_ZD_XY}}
    \hspace{0pt}
    \subfloat[]{\includegraphics[height=1.5in]{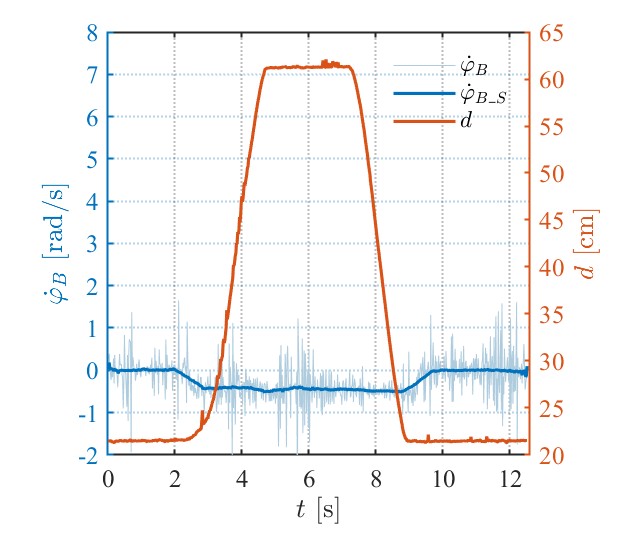}
    \label{fig_experiment_track_ZD_d}}
    \caption{Effect of $d$ variation on trajectory and speed. (a) and (b) are of mobile along $B_x$. (c) and (d) are of mobile along $B_y$. (e) and (f) are of mobile along $B_x$ and around $B_z$. The starting coordinates are (0,0).}
   \label{fig_experiment_track_D}
\end{figure}

To validate the impact of varying wheel spacing $d$ on movement during operation, three experiments were designed. These experiments involved changing the wheel spacing $d$ while moving in a straight line along the $B_x$ axis, in a straight line along the $B_y$ axis, and rotating around the $B_z$ axis. An OptiTrack motion capture system recorded the positions of the left wheel group center $L$, the right wheel group center $R$, and the prototype center $B$. Using the positional data from these experiments, the wheel spacing $d$ and the movement speed of the prototype center were calculated, as shown in Fig. \ref{fig_experiment_track_D}.

When moving along the $B_x$ axis and changing the wheel spacing $d$, the trajectories of the prototype center and the left and right wheel groups are shown in Fig. \ref{fig_experiment_track_D}\subref{fig_experiment_track_XD_XY}. The trajectory of the prototype center essentially follows a straight line along the X-axis. However, when the wheel spacing increases, the trajectory deviates. The primary reason for this deviation is the need to overcome greater static friction in the linear slides when extending the wheel spacing, as well as acceleration, which causes wheel slippage and generates additional torque, leading to extra angular velocity. Fig. \ref{fig_experiment_track_D}\subref{fig_experiment_track_XD_d} shows that $\dot{x}_B$ is minimally affected by changes in $d$.

When moving along the $B_y$ axis and changing the wheel spacing $d$, the trajectory of the prototype center is shown in Fig. \ref{fig_experiment_track_D}\subref{fig_experiment_track_YD_XY}. The trajectory of the prototype center exhibits significant errors along the Y-axis. The main reason is that dynamic balancing in the $B_x$ direction induces speed and displacement in the $B_x$ direction, particularly when $d$ changes, affecting the dynamic balance. In Fig. \ref{fig_experiment_track_D}\subref{fig_experiment_track_YD_d}, it is evident that the speed $\dot{y}_B$ is significantly influenced by changes in $d$.

When movement along the $B_x$ axis and rotation around the $B_z$ axis occur simultaneously, changing the wheel spacing $d$, the trajectory of the prototype center is shown in Fig. \ref{fig_experiment_track_D}\subref{fig_experiment_track_ZD_XY}. Initially, only $\dot{x}_B$ is present; then $\dot{\varphi}_B$ is added; finally, $\dot{\varphi}_B$ is removed, leaving only $\dot{x}_B$. The trajectory of the prototype center is minimally affected by changes in $d$. Fig. \ref{fig_experiment_track_D}\subref{fig_experiment_track_ZD_d} demonstrates the predetermined changes in $\dot{\varphi}_B$, which are largely unaffected by changes in $d$.

\subsection{Demonstration of the  passability}

\begin{figure}[!t]
    \centering
    \subfloat[]{\includegraphics[width=3.4in]{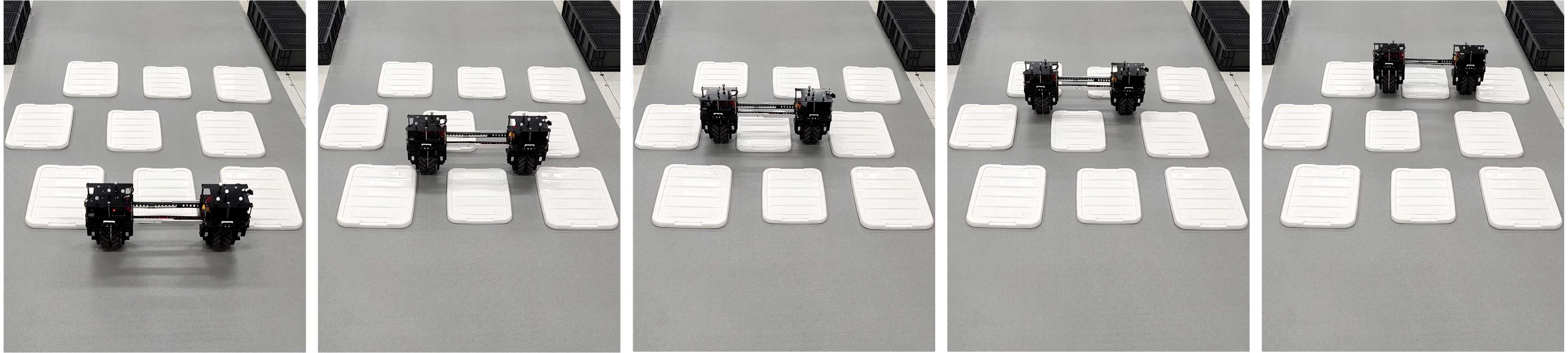}%
    \label{fig_demo_1}}
    \hspace{0pt}
    \subfloat[]{\includegraphics[width=3.4in]{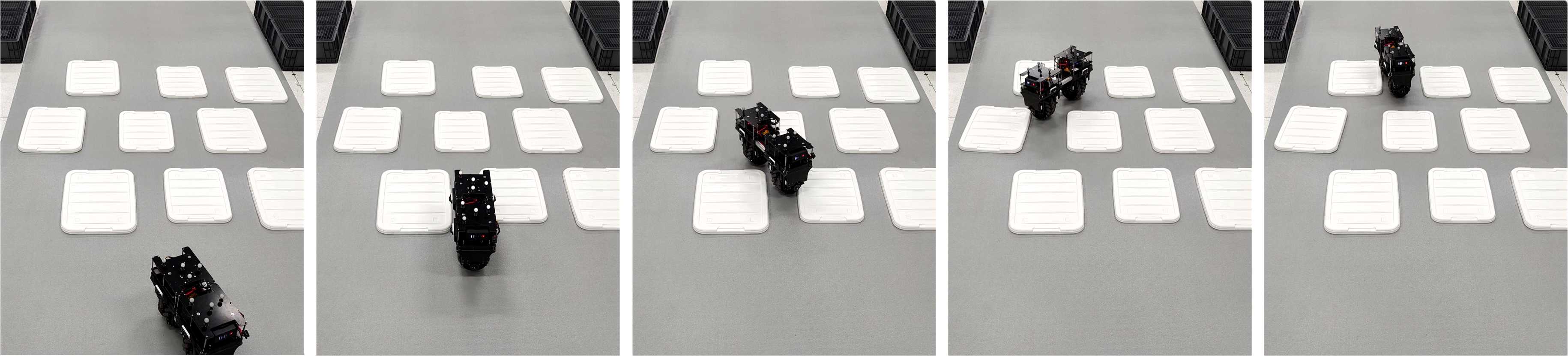}%
    \label{fig_demo_2}}
    \hspace{0pt}
    \subfloat[]{\includegraphics[width=3.4in]{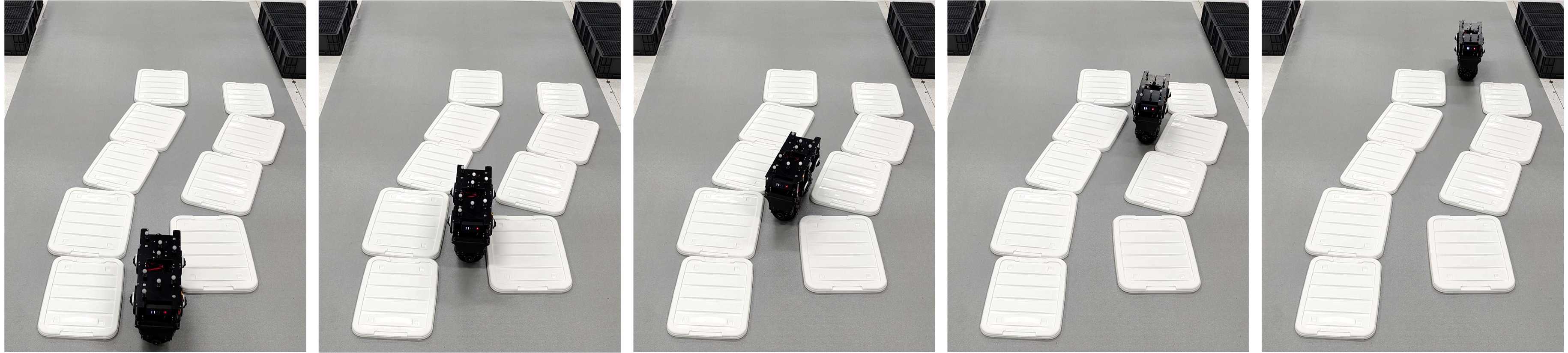}%
    \label{fig_demo_3}}
    \hspace{0pt}
    \subfloat[]{\includegraphics[width=3.4in]{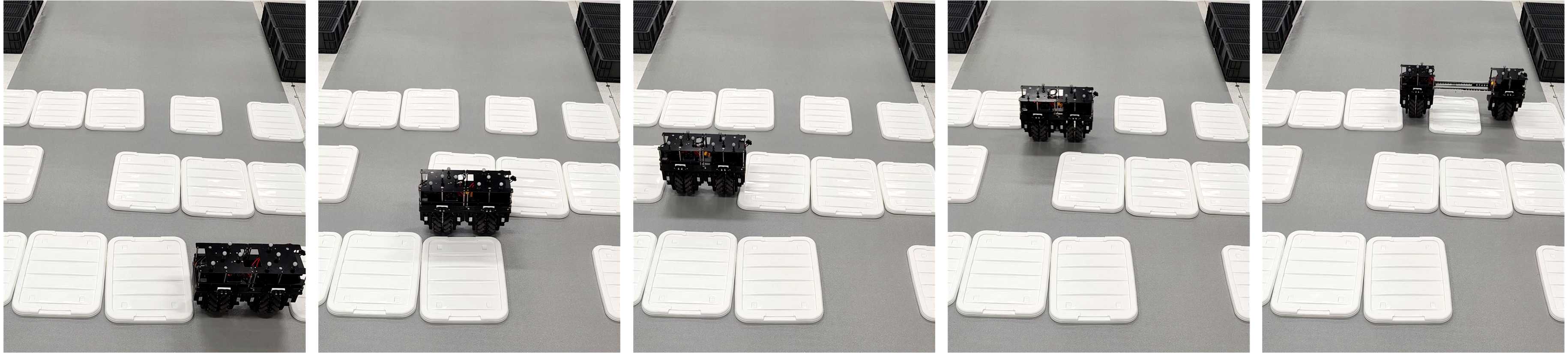}%
    \label{fig_demo_4}}    
    \caption{Demonstration of the passability.(a) Fixed orientation. (b) Lateral movement. (c) Narrow passage. (d) Changing $d$ in the path.}
    \label{fig_demo}
\end{figure}

To demonstrate the traversability of the proposed prototype, the following experiments were designed. Fig. \ref{fig_demo} shows images extracted from the demonstration video. Fig. \ref{fig_demo}\subref{fig_demo_1} illustrates the prototype navigating through a complex obstacle environment with an appropriate wheel distance, maintaining a constant orientation. Fig. \ref{fig_demo}\subref{fig_demo_4} builds upon this by adjusting the wheel distance during obstacle traversal to adapt to different channels. Fig. \ref{fig_demo}\subref{fig_demo_2} shows the prototype navigating the same obstacles as in Fig. \ref{fig_demo}\subref{fig_demo_1} by combining omnidirectional movement and varying wheel distance. Due to self-balancing control in the $B_x$ direction, the prototype oscillates back and forth to maintain balance, making control more challenging. Fig. \ref{fig_demo}\subref{fig_demo_3} demonstrates the use of omnidirectional movement to pass through a narrow passage.
 
\section{Conclusion}
In this study, we developed a novel Omni Differential Drive (ODD) wheeled mobility inspired by human movements and designed a prototype based on collinear Mecanum wheels to implement and verify the ODD. The ODD can achieve simultaneous reconfiguration and omnidirectional mobility for wheeled robots, meeting the requirements for mobility, agility, and stability in human living environments. Moreover, the kinematics of the ODD were modeled, and this model was used to establish the kinematic model of the prototype. A controller based on Parallel Cascade PID was designed to control the prototype's movement, and the kinematic models of both the ODD and the prototype were experimentally validated.

For future work, given the unique characteristics of the ODD, it will be applied to more chassis configurations. For instance, by adding castors, it can be transformed into a multi-wheeled vehicle that is reconfigurable and omnidirectionally mobile without requiring dynamic balancing. Additionally, it can be applied to wheel-legged robots where the omnidirectional movement and adjustable spacing of the wheel groups on both sides can cooperate with the leg joints to perform more complex actions. Furthermore, we will design lighter and more compact active omnidirectional wheels and use the ODD model to drive them. Path planning for mobile robots based on the ODD model will also be a focus of future research. In addition to future work on the ODD model, improving the control method of the prototype is also important. Currently, it is controlled using a kinematic model, which generates additional torques, so considering a dynamic model for control will be explored.

\balance
\bibliographystyle{IEEEtran}

\end{document}